%% file: main.tex
\documentclass[]{youtu} 

\usepackage{mathpazo}
\usepackage{graphicx}
\usepackage{natbib}
\usepackage{amsmath}
\usepackage{algorithm}
\usepackage{algpseudocode}
\usepackage{amssymb}
\usepackage{mathtools}
\usepackage{amsthm}

\usepackage{wrapfig}
\usepackage{enumitem}
\usepackage[table]{xcolor} 
\usepackage{booktabs}
\definecolor{deepblue}{RGB}{0, 0, 139}
\hypersetup{
    colorlinks=true,
    linkcolor=black,
    citecolor=deepblue,
    urlcolor=deepblue
}

\definecolor{lightpink}{rgb}{1.0, 0.95, 0.95}

\newcommand{\minisection}[1]{\vspace{5pt}\noindent\textbf{#1.}}

\title{ReMiT: RL-Guided Mid-Training for Iterative \\ LLM Evolution}
\author{Junjie Huang\textsuperscript{$1 \dagger$}, Jiarui Qin\textsuperscript{$2 \dagger$}, Di Yin\textsuperscript{$2$}, Weiwen Liu\textsuperscript{$1 \ddagger$}, Yong Yu\textsuperscript{$1$}, \\ Xing Sun\textsuperscript{$2$}, Weinan Zhang\textsuperscript{$1 \ddagger$}}
\affiliation{$^1$Shanghai Jiao Tong University\\\textsuperscript{$2$}Tencent Youtu Lab}

\youtufinalcopy 

\begin{document}

\abstract{
Standard training pipelines for large language models (LLMs) are typically unidirectional, progressing from pre-training to post-training. However, the potential for a bidirectional process—where insights from post-training retroactively improve the pre-trained foundation—remains unexplored. We aim to establish a \textbf{self-reinforcing flywheel}: a cycle in which reinforcement learning (RL)-tuned model strengthens the base model, which in turn enhances subsequent post-training performance, requiring no specially trained teacher or reference model.
To realize this, we analyze training dynamics and identify the mid-training (annealing) phase as a critical turning point for model capabilities. This phase typically occurs at the end of pre-training, utilizing high-quality corpora under a rapidly decaying learning rate. 
Building upon this insight, we introduce \textbf{ReMiT} (\textbf{Re}inforcement Learning-Guided \textbf{Mi}d-\textbf{T}raining).
Specifically, ReMiT leverages the reasoning priors of RL-tuned models to dynamically reweight tokens during the mid-training phase, prioritizing those pivotal for reasoning. Empirically, ReMiT achieves an average improvement of 3\% on 10 pre-training benchmarks, spanning math, code, and general reasoning, and sustains these gains by over 2\% throughout the post-training pipeline. These results validate an iterative feedback loop, enabling continuous and self-reinforcing evolution of LLMs.
}
\maketitle

\begin{figure}[htbp]
    \centering
    \includegraphics[width=0.97\linewidth]{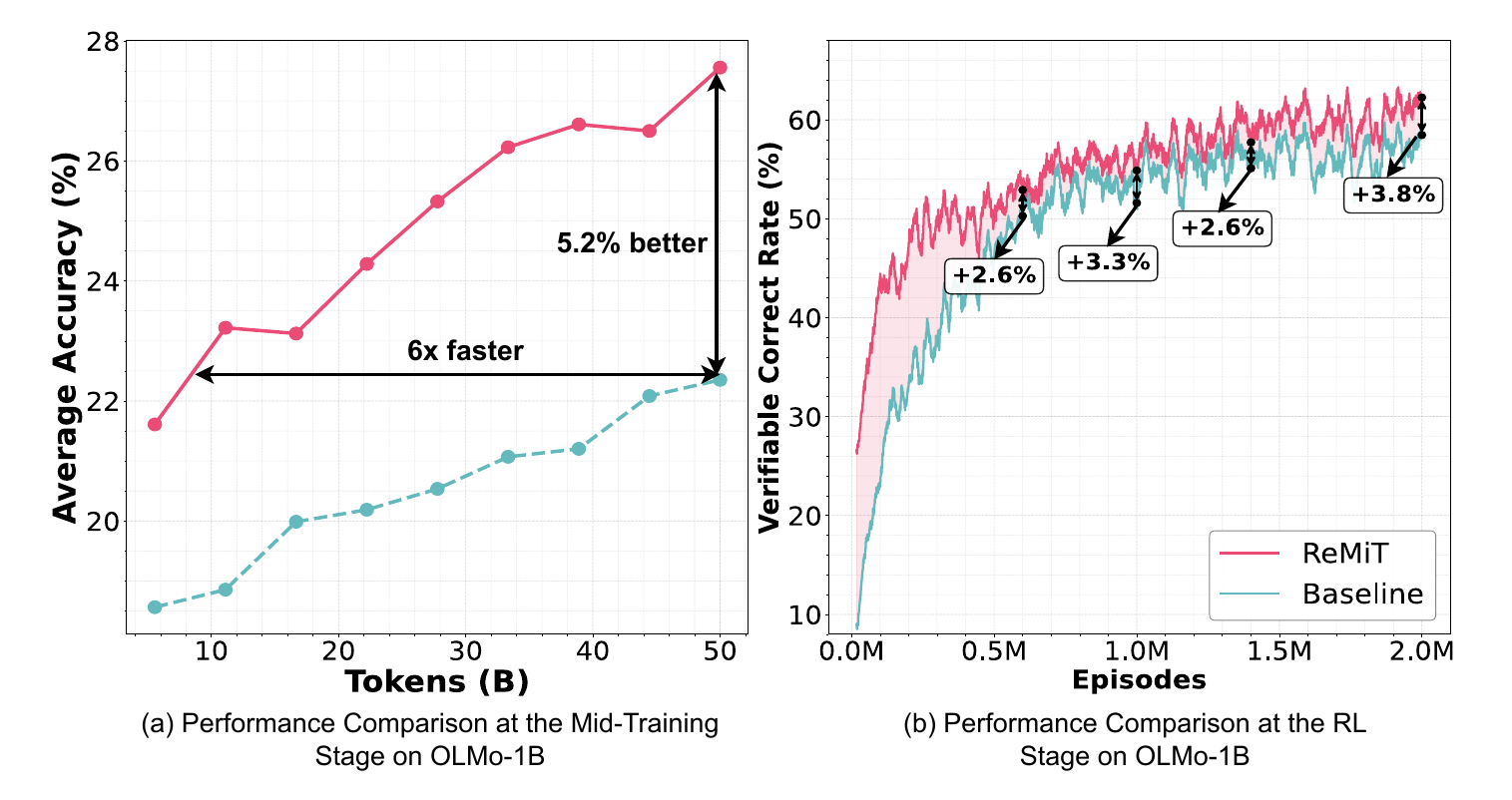}
    \caption{ReMiT on OLMo-1B substantially outperforms baselines trained with the standard mid-training method. (a) At the mid-training stage, ReMiT improves average accuracy across 10 widely-used benchmark tasks by 5.2\% and reaches the baseline performance $6\times$ faster. (b) The improvements can carry over to post-training: during RL, ReMiT maintains a higher verifiable correct rate than the baseline and achieves better performance.}
    \label{fig:abstract}
\end{figure}

\renewcommand{\thefootnote}{}
\footnotetext{$\dagger$ Equal contribution. huangjunjie2019@sjtu.edu.cn, jaredqin@tencent.com}
\footnotetext{$\ddagger$ Corresponding author.}

\input{paper/introduction}
\input{paper/preliminaries}

\input{paper/method}

\input{paper/theory}
\input{paper/experiments}
\input{paper/related_work}
\input{paper/conclusion}



\setcitestyle{numbers,square}
\setcitestyle{square,numbers,comma}
\bibliography{youtu_bib}

\input{paper/appendix}

\end{document}

%% file: paper/introduction.tex
\section{Introduction}\label{sec:intro}


Training large language models (LLMs) is a multi-stage process comprising pre-training and post-training
\citep{achiam2023gpt,touvron2023llama,yang2025qwen3}.
Pre-training of LLMs typically optimizes a next-token prediction objective that applies a uniform weight to every token in the loss function. In contrast, reinforcement learning (RL) follows a different training paradigm: RL methods assign non-uniform token weights based on reward signals from the environment~\citep{schulman2017proximal,shao2024deepseekmath}, yielding substantial improvements in the reasoning capabilities.
A strong pre-trained base model serves as the foundation for effective post-training. 
However, little research examines how to leverage these two stages synergistically, specifically, how post-trained models can provide guidance for pre-training, thereby improving final performance.

\begin{figure*}[!t]
    \centering
    \includegraphics[width=\linewidth]{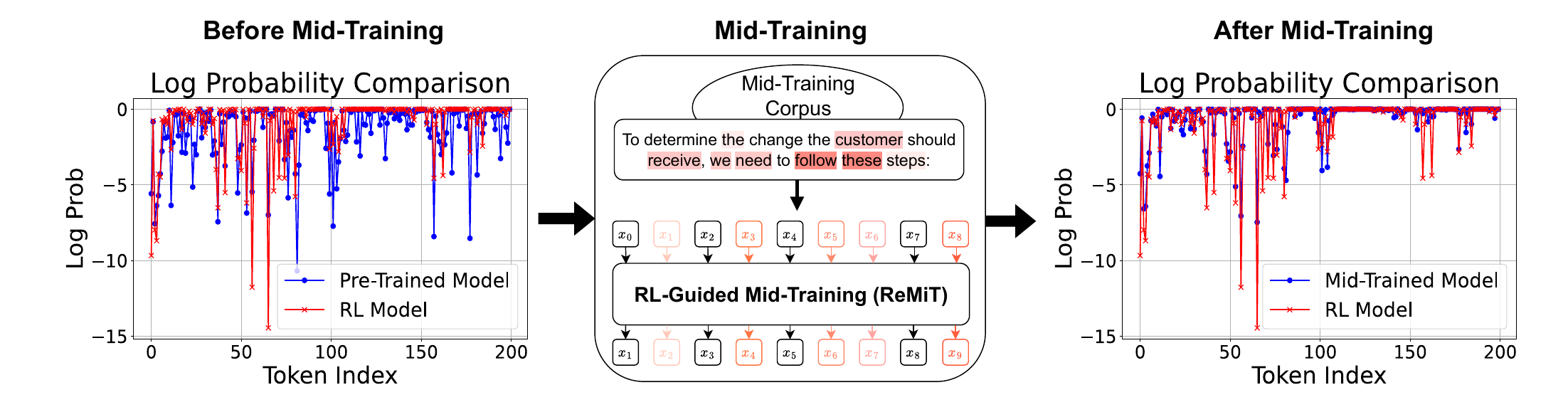}
    \caption{We identify the mid-training stage as a critical turning point, as it rapidly shifts the base model’s token distribution toward that of a more capable RL model. ReMiT enhances this stage by dynamically reweighting tokens in the mid-training corpus.}
    \label{fig:illustration}
\end{figure*}
Much of the literature on improving pre-training focuses on data selection, which can be viewed as document-level reweighting. 
Methods in this line of work~\citep{wenzek2019ccnet,xie2023doremi,albalak2024survey} primarily perform data cleaning to quickly filter noisy documents. 
Other studies investigate sample-level reweighting~\citep{gu2023minillm,zhang2025survey}, but these approaches often rely on heuristic rules and remain relatively coarse-grained. 
Recent research suggests that only a small subset of tokens is pivotal for model capability~\citep{arbuzov2025beyond, cui2025entropy}, indicating that sentence-level upsampling or downsampling may be suboptimal and motivating finer-grained token-level reweighting.

By analyzing training dynamics during the pre-training phase, we identify the mid-training stage as a critical turning point (see Appendix~\ref{sec:app:bg} for details).
This stage typically trains on the highest-quality corpora under a rapidly decaying learning rate to accelerate the acquisition of high-order capabilities such as graduate-level STEM knowledge or mathematical/code reasoning~\citep{touvron2023llama,yang2025qwen3}.
Our analysis indicates that mid-training induces a significant shift in the probability distribution of the base model.
As shown in Figure~\ref{fig:illustration}, we compare the token-level log probability distributions over sequences produced by the final RL model with those of the base model before and after mid-training. 
It can be observed that the RL model is much closer to the mid-trained model, while it differs significantly from the pre-trained model.
The observation suggests that mid-training plays a pivotal role in the qualitative transformation of the base model.
Since the RL model and the mid-trained model exhibit similar token-level probability distributions, it is natural to consider whether the RL model can be used to refine the token loss distribution during mid-training. 
By doing so, we aim to produce a model that is more closely aligned with the RL model, thereby yielding a stronger base. This improved base could be more compatible with the subsequent RL phase and ultimately lead to better overall performance.


Building on these observations, we propose ReMiT (\textbf{Re}inforcement Learning-Guided \textbf{Mi}d-\textbf{T}raining). ReMiT bridges pre-training and post-training by reusing the in-pipeline RL-tuned model as a reference, thereby eliminating the need for a separately trained teacher model as in~\citet{lin2024not}. Specifically, we compute the per-token loss discrepancy between the RL reference and the base model to derive dynamic importance weights. Rather than discarding tokens via hard selection, ReMiT dynamically modulates their weights, thereby preserving the semantic coherence of the training data. Consequently, the mid-training stage effectively mimics the RL training paradigm by prioritizing tokens that contribute most to downstream performance, while largely retaining the efficiency of standard next-token prediction within the existing pipeline.



In summary, the contributions of this paper are as follows:
\begin{itemize}[leftmargin=*, topsep=5pt, partopsep=0pt, itemsep=0pt, parsep=11pt] 
    \item \textbf{Novel Mid-Training Framework:} We propose ReMiT, the first framework to leverage an RL-tuned model to guide \textbf{token-level weight assignment during the mid-training phase}. By dynamically prioritizing tokens pivotal for reasoning—in contrast to standard uniform weighting—ReMiT effectively transfers advanced capabilities retroactively to the base model.
    \item \textbf{Bidirectional Synergy without External Teachers:} We establish a bidirectional synergy between pre-training and post-training by reusing in-pipeline models. Notably, ReMiT eliminates the need for specially trained or external reference models, ensuring practicality and scalability.
    \item \textbf{Empirical Validation and Flywheel Effect:} Extensive experiments across diverse model families demonstrate the efficacy of ReMiT, yielding an average improvement of 3\% on 10 pre-training benchmarks. Crucially, we verify that these gains persist through post-training, sustaining an improvement of over 2\% throughout the  pipeline, thereby establishing a co-improving flywheel between the base and RL models for continuous evolution.
\end{itemize}


%% file: paper/preliminaries.tex
\section{Preliminaries and Empirical Motivation}
\subsection{Standard Mid-Training Formulation}
Standard training pipelines for LLMs typically conclude the pre-training phase with a specialized stage known as mid-training (see Appendix~\ref{sec:app:bg}). This stage focuses on refining the model using high-quality corpora (e.g., mathematical reasoning or code) under a rapidly decaying learning rate. Despite the shift in data distribution, the fundamental objective remains the standard Next-Token Prediction (NTP).

Consider a sequence of tokens, $x_{1:T}=[x_1,\dots,x_T]$ from a vocabulary $\mathcal{V}$. An autoregressive model defines the probability of this sequence as the product of conditional probabilities, as shown in Equation~\eqref{equ:ntp1}.
Given a mid-training dataset $\mathcal{D}_{\text{mid}}$ comprising $N$ sequences $\mathcal{D}_{\text{mid}}=\{x^{(i)}_{1:T_i}\}_{i=1}^{N}$, the model parameters $\theta$ are optimized by minimizing the average negative log-likelihood in Equation~\eqref{equ:ntp2}:
\begin{equation}\label{equ:ntp1}
p_\theta(x_{1:T}) \;=\; \prod_{t=1}^{T} p_\theta\!\left(x_t \mid x_{<t}\right).
\end{equation}
\begin{equation}\label{equ:ntp2}
\mathcal L_{\text{NTP}}(\theta)= -\,\frac{1}{\sum_i T_i}\sum_{i=1}^{N}\sum_{t=1}^{T_i}
\log p_\theta\!\left(x^{(i)}_{t}\mid x^{(i)}_{<t}\right). 
\end{equation}

This standard objective applies a uniform weighting to every token position $t$. Such indiscriminate treatment fails to distinguish between low-entropy tokens and pivotal tokens that drive complex reasoning steps, a limitation that motivates our proposed framework.

\subsection{Analyzing Distributional Discrepancies Between Base and RL Models}\label{sec:bg2}
We visualize the distributional shift by comparing the token-level log-probability gap between the base and RL models in Figure~\ref{fig:illustration}. Specifically, we compute probability distributions over sequences drawn from the mid-training corpus. We observe that the gap narrows substantially after mid-training, confirming that this stage acts as a critical turning point, driving a qualitative transformation in the base model.

\begin{figure}[!t]
    \centering
    \includegraphics[width=\linewidth]{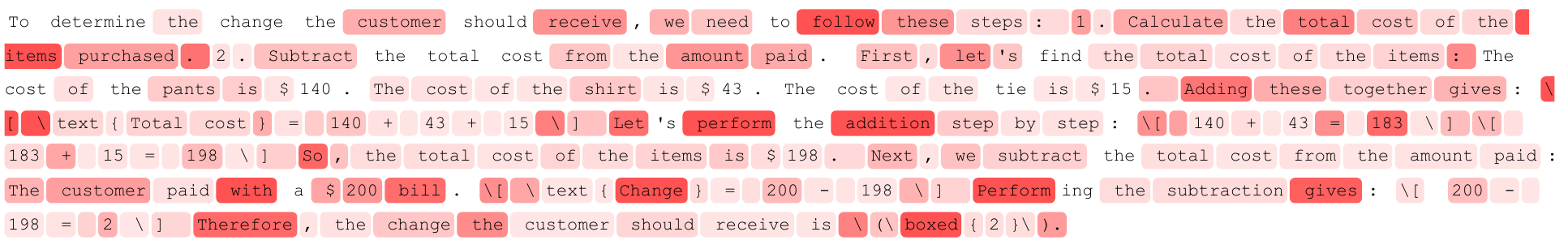}
    \caption{Visualization of the log-probability divergence between the pre-trained base model and the RL model. Background intensity reflects the margin $\Delta\log p=\log p_{\mathrm{RL}}-\log p_{\mathrm{base}}$, where deeper red highlights pivotal tokens on which the RL model demonstrates significantly higher confidence than the base model.}
    \label{fig:visualization}
\end{figure}
To pinpoint exactly where the RL model outperforms the pre-trained base model before mid-training, Figure~\ref{fig:visualization} visualizes the per-token prediction difference ($\Delta\log p$). The background color encodes this margin: deeper red indicates that the RL model assigns significantly higher likelihood to the ground-truth token than the base model.
Visually, we observe that the advantage of RL model is highly localized. While the majority of the text retains a light background, reflecting modest margins, distinct pivotal tokens emerge with deep red highlighting.
This pattern suggests that the superior reasoning of the RL model is driven by a small set of pivotal tokens, with the rest largely following the distribution of the base model. Notably, these high-margin tokens often correspond to discourse connectives (e.g., "Therefore", "So"), structural markers (e.g.,  "boxed"), or key logical verbs (e.g., "follow", "Adding"), consistent with the enhanced reasoning capabilities of RL models. See Appendix~\ref{sec:app:visualize} for more examples.

Building on these observations, we propose to dynamically upweight these pivotal tokens. We identify the mid-training phase as the optimal window for this intervention. Unlike general pre-training, mid-training utilizes high-quality corpora—enriched with instruction and reasoning-intensive data—that are inherently aligned with post-training objectives. Furthermore, due to the rapidly decaying learning rate, this phase serves as the critical stage for acquiring high-order capabilities. 

%% file: paper/method.tex
\section{The ReMiT Framework}\label{sec:method}
\subsection{Framework Overview}
\begin{figure*}[!t]
    \centering
    \includegraphics[width=0.88\linewidth]{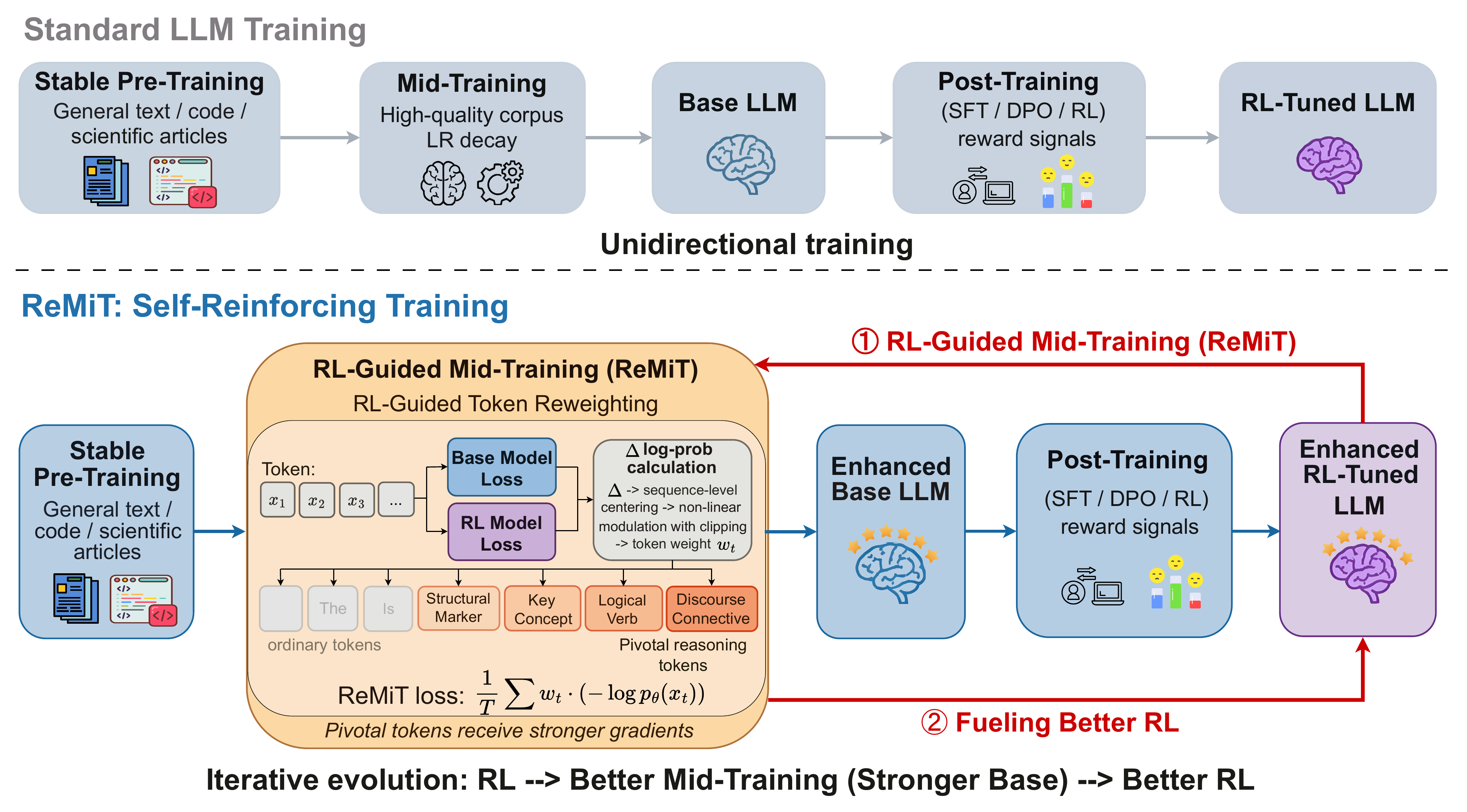}
    \caption{Overview of the proposed ReMiT framework. The pipeline connects pre-training and post-training, establishing a self-reinforcing flywheel: improvements from the RL stage are retroactively transferred to strengthen the base model foundation, which in turn amplifies performance in subsequent post-training stages.}
    \label{fig:pipeline}
\end{figure*}



As illustrated in Figure~\ref{fig:pipeline}, ReMiT introduces a paradigm shift from standard unidirectional training pipelines to a bidirectional, self-reinforcing flywheel. By retroactively transferring reasoning priors from the post-training stage back to the pre-training foundation, ReMiT creates a closed-loop system: utilizing a more capable RL reference for mid-training yields a strengthened base model, which in turn unlocks potential for subsequent post-training stages.

ReMiT is designed with two core principles:

\begin{itemize}[leftmargin=*, topsep=1pt, partopsep=0pt, itemsep=0pt, parsep=3pt] 
    \item \textbf{Efficient Reuse of In-Pipeline Models:} ReMiT directly repurposes the RL-tuned model inherently produced by the previous training cycle as the reference. This eliminates the need for training separate teacher models on manually curated clean datasets—a process that is often costly and ill-defined.
    \item \textbf{Strategic Intervention at Mid-Training:} We identify mid-training as the optimal phase due to its reasoning-oriented corpora and rapid learning rate decay. This data-efficient environment enables the base model to rapidly assimilate the reasoning priors from the RL reference.
\end{itemize}

\subsection{Dynamic Token Reweighting Mechanism}
ReMiT introduces a dynamic weighting scheme designed to prioritize tokens where the base model significantly underperforms relative to the RL reference. This approach effectively steers the optimization focus toward pivotal tokens that embody the RL model's reasoning priors.

\minisection{Quantifying Discrepancy}
We focus on tokens where the RL model exhibits higher confidence than the current base model. We quantify this discrepancy as the \textit{delta loss} $L_{\Delta}(x_t)$, defined as the difference in log-likelihoods:
\begin{equation}\label{equ:delta_loss}
\begin{split}
    L_{\Delta}(x_t) &= L_\theta(x_t)-L_{\mathrm{RL}}(x_t) \\
    &= -\log p_\theta(x_t\mid x_{<t})+\log p_{\text{RL}}(x_t\mid x_{<t}).
\end{split}
\end{equation}

\minisection{Sequence-Level Centering}
Raw log-probability gaps can vary significantly across different sequences due to intrinsic sentence difficulty. Sequence-level centering is used as a normalization mechanism to control per-sequence scale variation, ensuring stable and comparable token-level weights: \begin{equation}\label{equ:normalization}
\mu_{\Delta}=\frac{1}{T}\sum_{t=1}^{T} L_{\Delta}(x_t), \quad
\widehat{L}_{\Delta}(x_t)=L_{\Delta}(x_t)-\mu_\Delta.
\end{equation}

\minisection{Non-Linear Modulation with Clipping}
To map these centered deltas into scalar weights, we employ a scaled sigmoid function. This mapping is constructed to satisfy a neutrality constraint: when the two models are equally confident ($\widehat{L}_{\Delta}(x_t) \approx 0$), the objective should reduce to the standard NTP loss. Since $\sigma(0)=0.5$, a scaling factor of 2 is therefore required to ensure $w_t \approx 1$, thereby preserving the default learning signal.
To maintain training stability, we additionally enforce a clipping mechanism bounded by a hyperparameter $\epsilon$:
\begin{equation}\label{equ:weights}
w_t=\operatorname{clip}\!\Big(2\cdot\sigma\!\big(\widehat{L}_{\Delta}(x_t)\big),\,1-\epsilon,\,1+\epsilon\Big), \quad
\sigma(z)=\frac{1}{1+e^{-z}}.
\end{equation}

\begin{minipage}[t]{0.46\textwidth}
    \vspace{11pt} 
    This clipping mechanism serves as a regularization term or safety rail, ensuring that the objective remains anchored to the fundamental NTP paradigm. In particular, it prevents over-fitting to individual tokens and avoids degradation of basic syntactic coherence due to vanishing weights.
    \vspace{8.5pt}
    
    Together, this bounded monotonic mapping converts log-probability gaps into token weights in a controlled manner, enabling stable optimization. While other monotonic functions could be used, the scaled sigmoid offers desirable properties of smoothness and boundedness. We provide an ablation study on $\epsilon$ in Section~\ref{sec:ablation}.
\end{minipage}%
\hfill 
\begin{minipage}[t]{0.5\textwidth}
    \vspace{0pt} 
    \begin{algorithm}[H] 
        \caption{ReMiT: RL-Guided Mid-Training}
        \label{alg:rga}
        \begin{algorithmic}[1]
            \Require Params $\theta$, frozen RL ref, corpus $\mathcal{D}_{\text{mid}}$, clip $\epsilon$
            \For{$x_{1:T} \sim \mathcal{D}_{\text{mid}}$}
                \State $L_\theta(x_t) \gets -\log p_{\theta}(x_t \mid x_{<t})$
                \State $L_{\text{RL}}(x_t) \gets -\log p_{\text{RL}}(x_t \mid x_{<t})$
                \State $\widehat{L}_{\Delta}(x_t) \gets L_\theta(x_t) - L_{\text{RL}}(x_t) - \mu_{\Delta}$ 
            
                \Comment{Centered log-prob gap}
                \State $w_t \gets \operatorname{clip}\!\Big(2\sigma(\widehat{L}_{\Delta}),\,1-\epsilon,\,1+\epsilon\Big)$ 
                
                \Comment{Dynamic reweighting}
                \State $\mathcal{L}_{\text{ReMiT}} \gets \frac{1}{T}\sum_{t} w_t \cdot L_\theta(x_t)$
                \State $\theta \leftarrow \theta - \eta \nabla_\theta \mathcal{L}_{\text{ReMiT}}$
            \EndFor
        \end{algorithmic}
    \end{algorithm}
\end{minipage}

\minisection{The ReMiT Objective}
Finally, we integrate these dynamic weights into the standard objective. By performing soft reweighting rather than hard token selection, ReMiT preserves the semantic coherence of the sequence while amplifying the gradient signal for pivotal tokens: 
\begin{equation}\label{equ:rgaloss} 
\mathcal L_{\text{ReMiT}}(\theta)= -\,\frac{1}{\sum_i T_i}\sum_{i=1}^{N}\sum_{t=1}^{T_i} w^{(i)}_{t}
\log p_\theta\!\left(x^{(i)}_{t}\mid x^{(i)}_{<t}\right).
\end{equation}

The procedure is summarized in Algorithm~\ref{alg:rga}, which can be seamlessly integrated into existing pre-training pipelines. 


%% file: paper/theory.tex
\section{Theoretical Motivation}
\label{sec:theory}
Having introduced ReMiT in Section~\ref{sec:method}, we now analyze its theoretical foundations. A key question arises: \textit{why} does reweighting tokens based on an RL model effectively improve the base model? We address this by demonstrating that each ReMiT update actively steers the base model in a direction that locally reduces the Kullback-Leibler (KL) divergence to the KL-regularized optimal policy.

\subsection{ReMiT as Optimization towards an Implicit Target Distribution}
\label{subsec:implicit_target}

We first analyze the ReMiT objective for a fixed context $x_{<t}$. By viewing the weighted negative log-likelihood in Equation~\eqref{equ:rgaloss} through a distributional lens, we reveal that it is equivalent to minimizing the divergence towards a constructed \textit{implicit target distribution}. 
Let $w(x \mid x_{<t})$ denote the token-level weighting function derived from the log-probability gap.
We define the implicit target distribution $q_w$ as the re-weighted data distribution:
\begin{equation}
    q_w(x_t \mid x_{<t}) = \frac{1}{Z_w(x_{<t})} p_{\text{data}}(x_t \mid x_{<t}) w(x_t \mid x_{<t}),
\end{equation}
where $Z_w(x_{<t}) = \sum_{x' \in \mathcal{V}} p_{\text{data}}(x' \mid x_{<t}) w(x' \mid x_{<t})$ is the partition function ensuring normalization.
Intuitively, $q_w$ can be interpreted as an enhanced data distribution where pivotal tokens are assigned higher probability mass.

As derived in Appendix~\ref{app:decomposition}, the gradient of the ReMiT objective aligns with the gradient of the KL divergence between this implicit target and the current model:
\begin{equation}
    \scalebox{0.93}{$
    \nabla_\theta \mathcal{L}_{\text{ReMiT}}(\theta) \propto \nabla_\theta \mathbb{E}_{x_{<t} \sim \mathcal{D}} \left[ D_{\text{KL}}(q_w(\cdot\mid x_{<t}) \parallel \pi_\theta(\cdot\mid x_{<t})) \right].
    $}
\end{equation}
This decomposition reveals that ReMiT updates effectively drive the model distribution $\pi_\theta$ towards $q_w$. Through this process, the RL model influences optimization by \textbf{reshaping the implicit target}, thereby acting as a soft navigator for the base model.

\subsection{Directional Consistency with the Optimal Policy}
\label{subsec:consistency}

A critical question remains: does aligning with $q_w$ actually steer the model toward the true optimal policy $\pi^*$? To answer this, we reframe the mid-training process within the framework of KL-regularized reinforcement learning.

We model next-token generation as a decision step in a Markov decision process, where the context $x_{<t}$ serves as the state and the next token $x_{t}$ constitutes the action. We assume the existence of an underlying reward function $r(x_t, x_{<t})$ that captures the desired reasoning quality. The optimization objective is to maximize the expected reward while maintaining proximity to a reference policy $\pi_{\text{ref}}$ (typically the pre-trained base model) to prevent reward hacking:
\begin{equation}
    \max_{\pi} \mathbb{E}_{x_{<t} \sim \mathcal{D}, x_t \sim \pi} \left[ r(x_t, x_{<t}) - \beta \log \frac{\pi(x_t \mid x_{<t})}{\pi_{\text{ref}}(x_t \mid x_{<t})} \right],
\end{equation}
where $\beta > 0$ is the regularization coefficient.

\textbf{Local Geometry Analysis.} It is well-established in optimal control literature \citep{peters2010relative, rafailov2023direct} that the closed-form solution to this objective, $\pi^*$, follows a Gibbs distribution:
\begin{equation}
    \label{eq:optimal_policy}
    \pi^*(x_t \mid x_{<t}) \propto \pi_{\text{ref}}(x_t \mid x_{<t}) \exp \left( \frac{r(x_t, x_{<t})}{\beta} \right).
\end{equation}
With the explicit form of $\pi^*$ defined, we analyze the optimization landscape. Consider a local update of the current model $\pi_\theta$ in the direction of the implicit target $q_w$ derived in Section~\ref{subsec:implicit_target}. Let the updated policy be $\pi_{\theta'} = (1-\tau)\pi_\theta + \tau q_w$ for a sufficiently small step size $\tau > 0$.

A first-order expansion of the KL divergence $D_{\text{KL}}(\pi^* \parallel \pi_\theta)$ (derivation in Appendix~\ref{app:expansion}) demonstrates that the divergence from the optimal policy decreases provided that:
\begin{equation}
    \label{eq:consistency_condition}
    \sum_{x} \pi^*(x \mid x_{<t}) \left( \frac{q_w(x \mid x_{<t}) - \pi_\theta(x \mid x_{<t})}{\pi_\theta(x \mid x_{<t})} \right) > 0.
\end{equation}
This condition admits a clear interpretation: \textbf{if the implicit target $q_w$ assigns higher probability mass to tokens favored by $\pi^*$, the update is directionally consistent.}

Crucially, we leverage the RL model $\pi_{\text{RL}}$ as a specialized \textit{detector} for reasoning-critical tokens. Optimized for reasoning rewards, $\pi_{\text{RL}}$ exhibits significantly higher confidence than the base model specifically at pivotal decision points. Consequently, the log-probability gap functions as a reliable signal for identifying these tokens. By assigning $w_t > 1$ (and thus $q_w > \pi_\theta$) primarily when this gap is large, ReMiT actively concentrates optimization on tokens favored by $\pi^*$. This approach satisfies the consistency condition in Equation~\eqref{eq:consistency_condition} by utilizing $\pi_{\text{RL}}$ to scale the gradient magnitude on informative tokens, thereby effectively transferring reasoning priors while maintaining the base model's broader generative capabilities.

\subsection{Comparisons with Knowledge Distillation}
While ReMiT shares the high-level aspiration of transferring capabilities from the RL model to the base model, it fundamentally diverges from knowledge distillation (KD)~\cite{hinton2015distilling}. We characterize ReMiT as \textit{importance-aware discriminative distillation}, contrasting it with the \textit{mimicry} inherent in standard KD.

\minisection{Gradient Analysis}
To clarify the underlying difference, we analyze the gradients at a single token position $t$ rather than the aggregate loss in Equation~\eqref{equ:rgaloss}.

Standard KD~\cite{gou2021knowledge, yang2025survey} minimizes the forward KL divergence (equivalent to cross-entropy with soft targets) across the entire vocabulary $\mathcal{V}$:
\begin{equation}
\mathcal{L}_{\text{KD}} = - \sum_{x \in \mathcal{V}} \pi^*(x \mid x_{<t}) \log \pi_\theta(x \mid x_{<t}).
\end{equation}
Differentiating with respect to $\theta$ yields a gradient that compels the student to match the teacher's full probability distribution. Crucially, this formulation exerts gradient pressure even on low-probability tokens (often termed ``dark knowledge''~\cite{fan2024exploring, yu2024decoupling}).
\begin{equation}
\nabla_\theta \mathcal{L}_{\text{KD}} = - \sum_{x \in \mathcal{V}} \underbrace{\pi^*(x \mid x_{<t})}_{\text{Soft Target}} \nabla_\theta \log \pi_\theta(x \mid x_{<t}).
\end{equation}
In contrast, ReMiT operates on hard targets but dynamically modulates the learning intensity via a reweighted negative log-likelihood. The per-token gradient is:
\begin{equation}
\nabla_\theta L_{\text{ReMiT}}
= -\,w_t\cdot \nabla_\theta \log \pi_\theta(x_t \mid x_{<t}),
\end{equation}
where $w_t$ denotes the weight defined in Equation~\eqref{equ:weights}. 
Unlike KD, which alters the \textit{optimization direction} to match a potentially imperfect teacher $\pi_{\text{RL}}$, ReMiT preserves the ground-truth direction while utilizing $\pi_{\text{RL}}$ solely to scale the \textit{gradient magnitude} on informative tokens. 
We empirically compare ReMiT with KD in Section~\ref{sec:ablation} and Appendix~\ref{sec:app:exp}.


%% file: paper/experiments.tex
\section{Experiments}
\input{table/major_exp1}
\subsection{Experimental Setup}\label{sec:exp_setup}
\minisection{Base Models}
We evaluate ReMiT across three representative open-source model families that provide publicly available checkpoints suitable for mid-training: OLMo-1B~\citep{olmo20242}, SmolLM3-3B~\citep{smollm}, and Youtu-LLM-2B~\cite{lu2025youtu}.
In our experimental setup, the \textit{base model} is defined as the standard pre-trained checkpoint (the state immediately prior to mid-training), while the corresponding publicly released RL-tuned version serves as the reference policy.



\minisection{Baselines}
We compare ReMiT with four baselines:
\begin{itemize}[leftmargin=*, topsep=1pt, partopsep=0pt, itemsep=0pt, parsep=5pt]
    \item \textbf{Pre-Trained.} The standard pre-trained base model, which serves as the initialization point for mid-training.
    \item \textbf{Vanilla NTP.} The conventional next-token prediction objective that assigns uniform weights to all tokens during mid-training, as formulated in Equation~\eqref{equ:ntp2}.
    \item \textbf{MiniPLM~\citep{gu2024miniplm}.} 
    An offline sample selection framework that uses difference sampling to prioritize hard samples based on teacher probabilities. We adopt their offline selection strategy with a retention ratio of 0.8.
    \item \textbf{RHO-1~\citep{lin2024not}.} A token-level selection method that only retains tokens with high RHO loss. 
    Following their method, we mid-train the reference model on curated data and filter tokens with a top-$k$ retention ratio of 0.8.
\end{itemize}

\minisection{Implementation Details}
We conduct mid-training on 50B tokens using the official high-quality corpora associated with each model family.
For ReMiT, we set the clipping threshold $\epsilon=0.2$ in Equation~\eqref{equ:weights}.
To ensure rigorous comparison, we align all optimization hyperparameters (e.g., learning rates, batch sizes, sequence lengths) with the models' original pre-training configurations.
Detailed hyperparameters, corpus compositions, and training settings are strictly provided in Appendix~\ref{sec:app:eval}.

\minisection{Evaluation}
To comprehensively evaluate the effectiveness of ReMiT, we assess the few-shot performance of the resulting mid-trained models across multiple widely-used downstream benchmarks using the lm-eval-harness framework~\citep{gao2021framework}.
Our evaluation suite covers both reasoning-intensive domains, such as MATH and MBPP, and general capability benchmarks, including MMLU-Pro, TruthfulQA, and ARC-C.
To further demonstrate the downstream benefits of ReMiT, we also conduct subsequent post-training alignment—specifically SFT, DPO, and RLVR\footnote{\textbf{SFT}: supervised fine-tuning; \textbf{DPO}: direct preference optimization; \textbf{RLVR}: reinforcement learning with verifiable rewards.}—and evaluate the zero-shot accuracy of the resulting models.
Further evaluation protocols are detailed in Appendix~\ref{sec:app:eval}, with extended results provided in Appendix~\ref{sec:app:exp}.

\subsection{Main Results}
\begin{figure*}[!t]
    \centering
    \includegraphics[width=\linewidth]{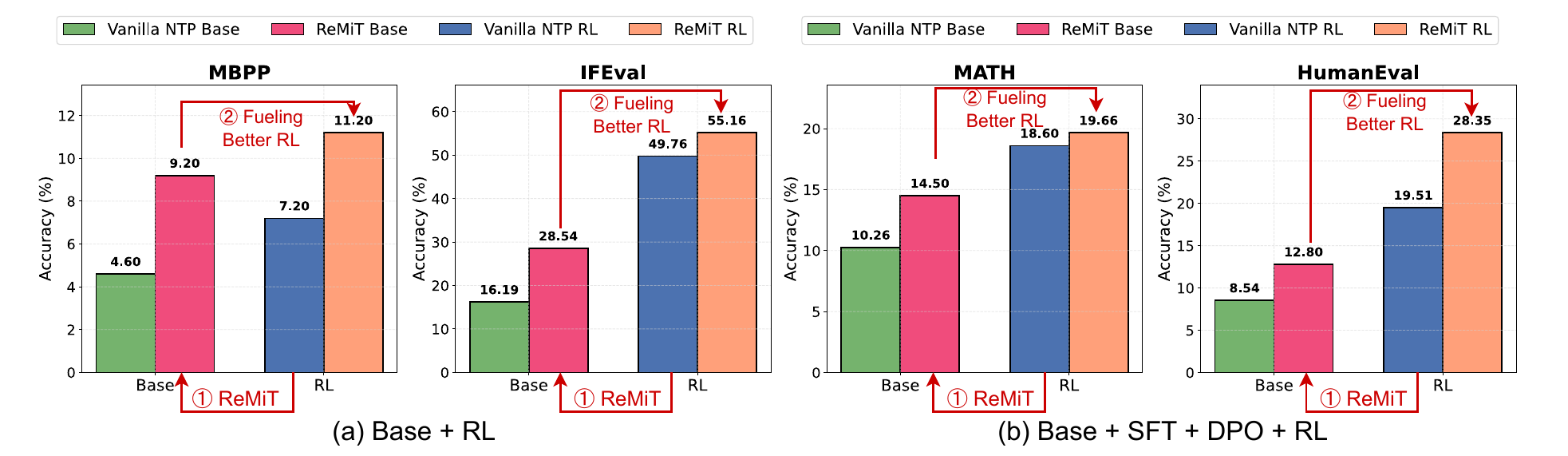}
    \vspace{-15pt}
    \caption{Performance gains of ReMiT on OLMo-1B acquired during mid-training transfer consistently to post-training, regardless of the post-training process (SFT, DPO, or RLVR). Figure (a): applying RL directly to the mid-trained base model. Figure (b): applying the complete post-training procedure to the mid-trained base model.}
    \label{fig:exp2}
\end{figure*}
\minisection{ReMiT Enhances Mid-Training Performance}
Table~\ref{tab:major_exp1} reports few-shot accuracies on downstream tasks, leading to three key observations. \textit{First}, mid-training proves to be a critical stage: all mid-trained variants significantly outperform the pre-trained base models. \textit{Second}, ReMiT achieves the best performance by dynamically prioritizing tokens pivotal to the advanced reasoning capabilities of the RL model. \textit{Finally}, ReMiT outperforms both RHO-1 and MiniPLM. Unlike RHO-1, which incurs costly auxiliary reference model training with uncertain gains and disrupts semantic coherence by discarding tokens, ReMiT employs a cost-free, soft reweighting strategy. Moreover, ReMiT offers finer granularity than MiniPLM's sample selection, ensuring critical informative tokens are not obscured. 

\begin{figure}[!t]
    \centering
    \begin{minipage}[t]{0.48\linewidth}
        \vspace{14pt}
        \centering
        \includegraphics[width=\linewidth]{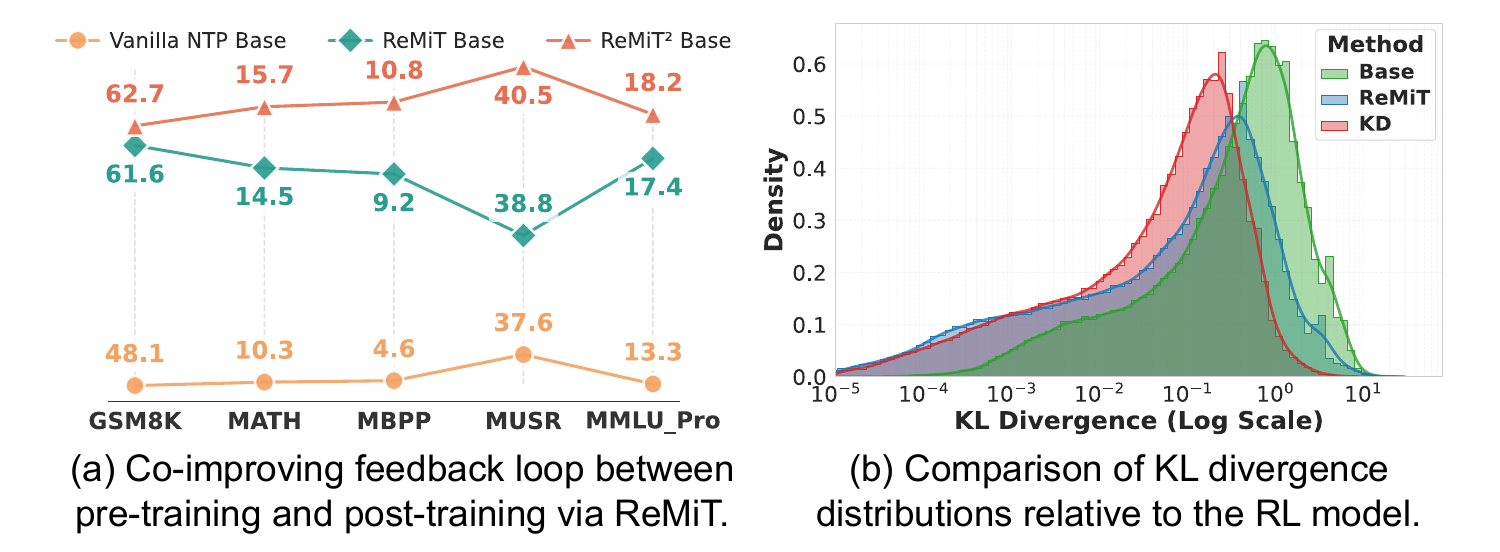}
        \caption{Robust generalization via ReMiT. (a) Co-improving Loop: Performance gains consistently transfer and amplify through iterative mid-training cycles (Vanilla NTP $\to$ ReMiT $\to$ ReMiT$^2$). (b) KL Analysis: While KD forces the model to strictly mirror the RL policy, ReMiT allows for a moderate KL divergence, ensuring better transferability during post-training.}
        \label{fig:exp3}
    \end{minipage}%
    \hfill 
    \begin{minipage}[t]{0.48\linewidth}
        \vspace{0pt}
        \centering
        \includegraphics[width=\linewidth]{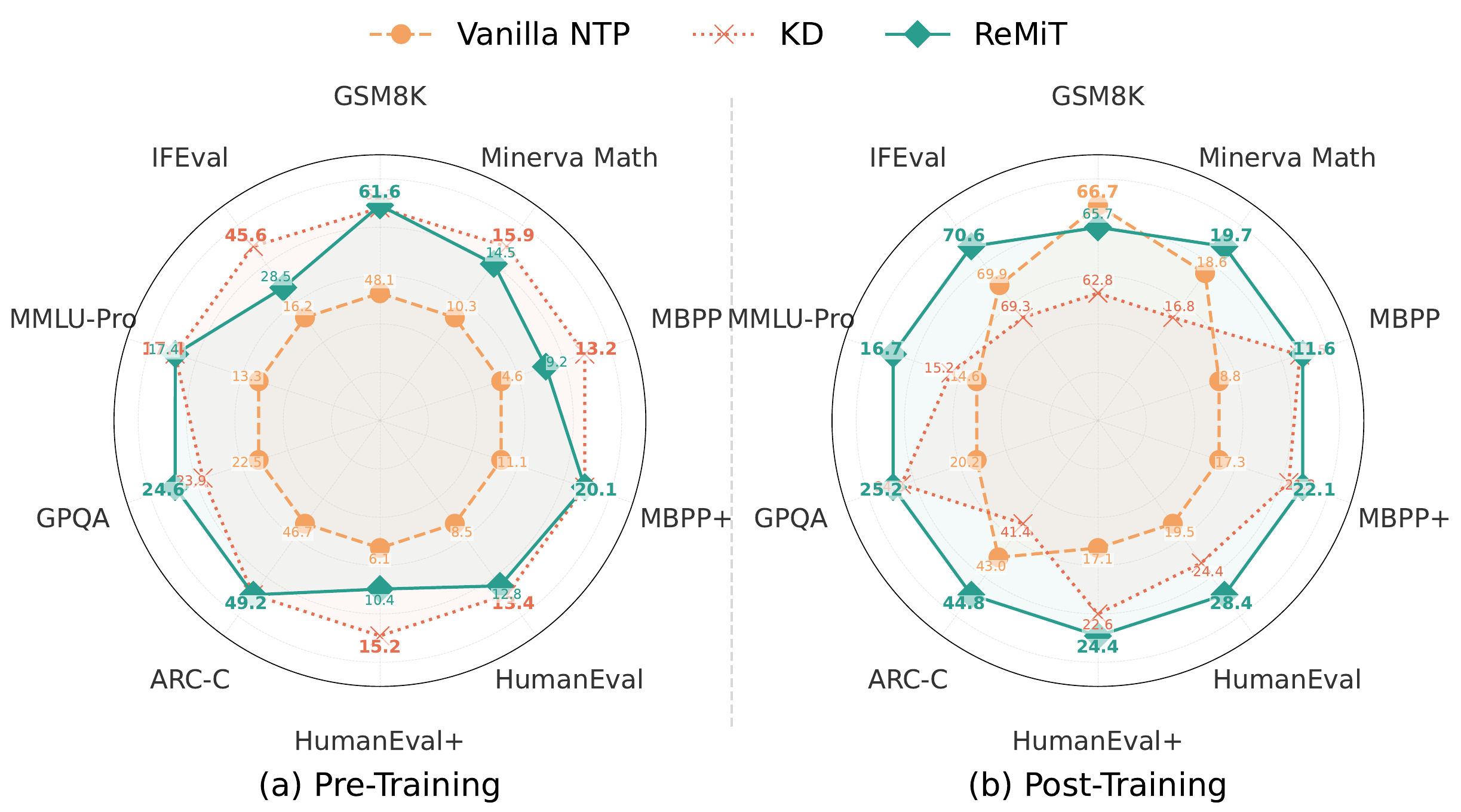}
        \caption{Sustained advantages of ReMiT across training stages. ReMiT achieves better generalization with pre-training gains that consistently carry over to post-training, whereas KD's early advantages fade in later stages.}
        \label{fig:radar}
    \end{minipage}
\end{figure}

\minisection{Mid-Training Gains Transfer to Post-Training}
A stronger base model \textit{does not necessarily guarantee} a superior final RL model after post-training, as further analyzed in Figure~\ref{fig:radar}.
To address this, we show that ReMiT enhances performance not only during the mid-training stage but also provides a superior foundation for subsequent post-training.
As illustrated in Figure~\ref{fig:exp2}, we apply identical post-training procedures to both the vanilla NTP model and ReMiT on OLMo-1B. 
We observe that \textit{the performance gains achieved during mid-training transfer consistently to post-training}, regardless of the specific technique employed. 
Figure~\ref{fig:exp2} (a) presents the results of applying RL directly to the mid-trained models, while Figure~\ref{fig:exp2} (b) shows the results after applying the full post-training pipeline, including SFT, DPO, and RL. Additional results are provided in Appendix~\ref{sec:app:exp}.

\minisection{Co-Improving Feedback Loop Between Pre-Training and Post-Training}
We have shown that ReMiT enhances performance during both pre-training and post-training. 
We further demonstrate that \textit{a stronger RL reference amplifies mid-training gains} (Figure~\ref{fig:exp3} (a)). By employing the initial ReMiT-derived RL model to guide a second round of mid-training, we produce ReMiT$^2$. This creates a cascading effect: an improved base yields a superior RL reference, which subsequently refines the base further. The performance superiority of ReMiT$^2$ over ReMiT confirms the framework's capability to drive continuous self-improvement.




\subsection{Analysis and Discussion}\label{sec:ablation}

\minisection{ReMiT vs. KD: Robust Generalization via Soft Alignment}
Our experiments reveal a critical distinction in how ReMiT and KD influence the model's long-term evolution. 
As illustrated in Figure~\ref{fig:radar}, while KD achieves impressive performance gains immediately after mid-training, these advantages fail to sustain after subsequent post-training. 
This diminishing return suggests that KD compromises the model's capacity for further adaptation.
We attribute this to the over-alignment phenomenon. 
In Figure~\ref{fig:exp3} (b), we analyze the KL divergence distributions relative to the teacher model. 
The KD model exhibits an extremely low KL divergence, indicating it has collapsed onto the teacher's specific distribution and over-fitted its output patterns.
In contrast, ReMiT maintains a moderate distributional gap. 
This soft alignment ensures the base model assimilates the teacher's high-level reasoning capabilities without strictly memorizing its probability distribution, thereby preserving the diversity and potential required for continuous self-improvement during post-training.
Extended results are provided in Appendix~\ref{sec:app:exp}.

\minisection{Choosing the Reference: Why RL over SFT}
To investigate the influence of reference policy selection, we conduct an ablation study comparing the RL model against an SFT baseline; for a detailed visualization and analysis of their distinct token-level preferences, please refer to Appendix~\ref{sec:app:visualize}.
Regarding downstream performance, Figure~\ref{fig:app:sga} demonstrates that ReMiT consistently outperforms the SFT baseline throughout the mid-training process.

\begin{minipage}[t]{0.48\textwidth}
    \vspace{8pt} 
    \minisection{Impact of the Clipping Mechanism}
    As analyzed in Figure~\ref{fig:clipping}, the clipping mechanism serves as a critical safety rail that prevents the gradient collapse observed in the `no clip' variant. By enforcing a baseline weight, clipping preserves general modeling capabilities while prioritizing reasoning-critical tokens. This balance ensures healthy gradient norms and faster convergence, contrasting sharply with RHO-1, where aggressive token discarding results in diminished gradients and impedes efficient optimization.
\end{minipage}%
\hfill 
\begin{minipage}[t]{0.48\textwidth}
    \vspace{0pt} 
    \centering
    \includegraphics[width=\linewidth]{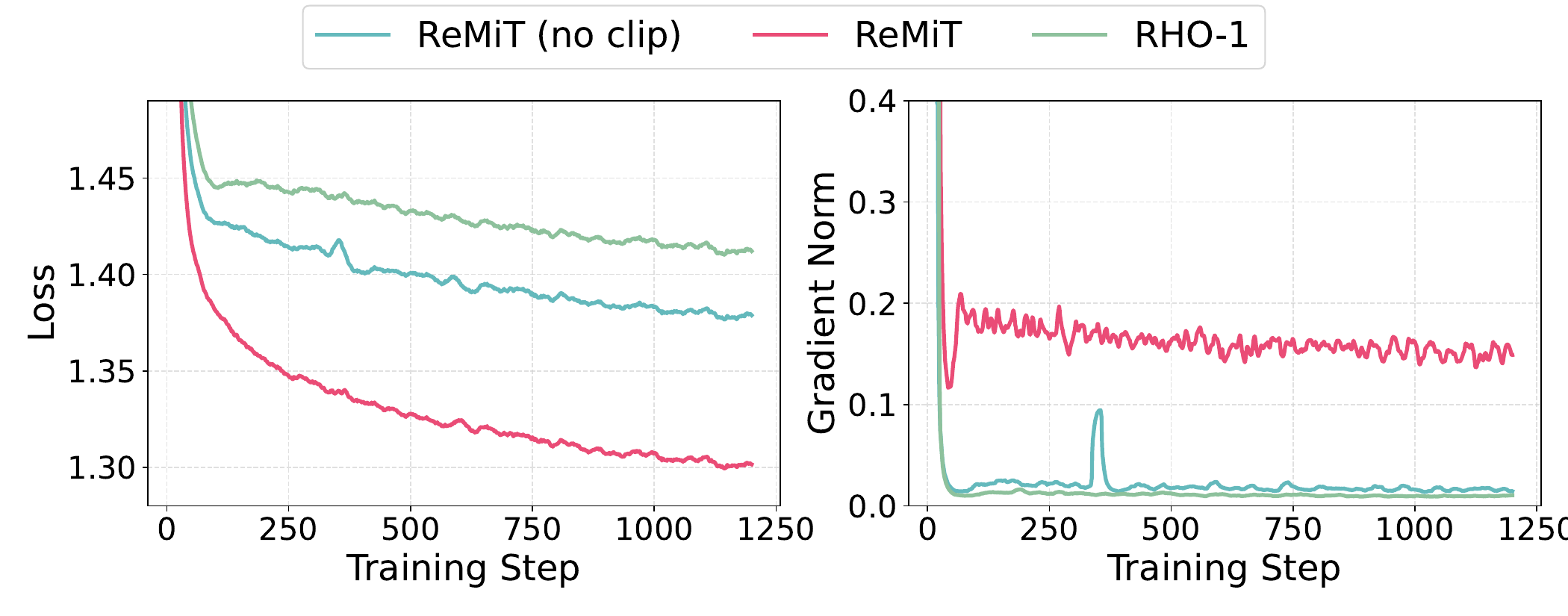}
    \captionof{figure}{Clipping mechanism in ReMiT on Youtu-LLM. By bounding token-level weights, clipping keeps gradient norms within a normal range and accelerates loss reduction.}
    \label{fig:clipping}
\end{minipage}





%% file: table/major_exp1.tex
\begin{table*}[!t]
\renewcommand\arraystretch{1.1}
\setlength{\tabcolsep}{6pt}
\centering
\caption{Few-shot accuracy across 10 widely used downstream tasks. All improvements 
of ReMiT over baselines are statistically significant. Extended results and the num\_shots are provided in Appendix~\ref{sec:app:eval}. The best scores of each model family are \textbf{boldfaced}.}
\resizebox{0.98\textwidth}{!}{
\footnotesize
\begin{tabular}{c|cccccccccc|c}
\toprule
\textbf{Method} & \textbf{GSM8K} & \textbf{MATH} & \textbf{GPQA} & \textbf{BBH} & \textbf{IFE} & \textbf{HE} & \textbf{MBPP} & \textbf{TQA} & \textbf{ARC-C} & \textbf{MMLU\textsubscript{P}} & \textbf{Avg.} \\
\midrule
\rowcolor{lightpink}\multicolumn{12}{c}{Mid-Training on OLMo-1B} \\
\midrule
Pre-Trained & 3.03 & 2.94 & 20.31 & 28.43 & 22.66 & 6.71 & 4.80 & 21.30 & 44.71 & 9.54 & 16.44 \\
Vanilla NTP & 48.14 & 10.26 & 22.54 & 30.87 & 16.19 & 8.54 & 4.60 & 22.40 & 46.67 & 13.31 & 22.35 \\
MiniPLM & 48.45 & 9.60 & 23.21 & 30.38 & 16.79 & 7.32 & 6.80 & 23.13 & 45.31 & 13.15 & 22.41 \\
RHO-1 & 50.42 & 10.32 & \textbf{25.45} & 29.33 & 19.06 & 6.71 & 6.20 & 23.38 & 46.42 & 13.68 & 23.10 \\
ReMiT & \textbf{61.64} & \textbf{14.50} & 24.55 & \textbf{32.07} & \textbf{28.54} & \textbf{12.80} & \textbf{9.20} & \textbf{25.58} & \textbf{49.23} & \textbf{17.44} & \textbf{27.56}\\
\midrule
\rowcolor{lightpink}\multicolumn{12}{c}{Mid-Training on SmolLM3-3B}
\\
\midrule
Pre-Trained & 31.61 & 14.52 & \textbf{27.68} & 43.13 & 22.06 & 25.61 & 37.40 & 30.23 & \textbf{56.31} & 23.24 & 31.18\\
Vanilla NTP & \textbf{64.22} & 31.64 & 26.34 & 56.50 & 43.29 & 28.05 & 47.60 & 29.74 & 53.24 & 30.68 & 41.13 \\
MiniPLM & 60.65 & 30.84 & 26.34 & 57.10 & 41.49 & 29.88 & 48.00 & 29.13 & 53.58 & 29.52 & 40.65 \\
RHO-1 & 62.93 & 28.70 & 26.56 & 55.32 & 38.73 & 31.10 & 43.80 & 28.52 & 54.10 & 28.19 & 39.80\\
ReMiT & 63.76 & \textbf{31.68} & 26.12 & \textbf{58.27} & \textbf{45.68} & \textbf{37.20} & \textbf{49.60} & \textbf{31.95} & 54.69 & \textbf{30.73} & \textbf{42.97}\\
\midrule
\rowcolor{lightpink}\multicolumn{12}{c}{Mid-Training on Youtu-LLM-2B}
\\
\midrule
Pre-Trained & 34.87 & 15.30 & 22.99 & 43.39 & 23.26 & 25.00 & 38.80 & \textbf{27.54} & 52.30 & 19.96 & 30.34 \\
Vanilla NTP & 49.51 & \textbf{25.00} & 27.90 & 44.37 & 32.61 & 37.20 & 46.60 & 26.56 & 53.33 & 24.93 & 36.80\\
MiniPLM & 51.86 & 22.76 & 26.34 & 44.20 & 33.09 & 35.98 & \textbf{47.00} & 26.56 & 53.75 & 25.34 & 36.69 \\
RHO-1 & 42.46 & 16.44 & 26.79 & 40.45 & 32.01 & 27.44 & 39.20 & 26.68 & 49.23 & 21.69 & 32.24\\
ReMiT & \textbf{52.69} &  24.50 & \textbf{29.69} & \textbf{47.21} & \textbf{36.93} & \textbf{39.94} & \textbf{47.00} & 27.42 & \textbf{54.61} & \textbf{25.85} & \textbf{38.58}\\
\bottomrule
\end{tabular}
}
\vspace{-4mm}
\label{tab:major_exp1}
\end{table*}

%% file: paper/related_work.tex
\section{RELATED WORK}

\minisection{Data Selection Strategies in Pre-Training}
Optimizing data quality is fundamental to pre-training~\citep{xie2023data, albalak2024survey, liu2024datasets, gu2024miniplm,zhu2025toremi}. 
Traditional strategies employ lightweight filters, including heuristic-based, classifier-based~\citep{mann2020language}, and perplexity or loss-based approaches~\citep{Qin2023InfoBatchLT,wenzek2019ccnet}, while recent approaches utilize reference models for hard data selection based on loss signals~\citep{Ankner2024PerplexedBP, lin2024not}. 
In contrast, ReMiT introduces a token-level \textit{reweighting} mechanism that leverages the existing RL model as a cost-free reference, avoiding the overhead of training auxiliary models.


\minisection{Token-Level Analysis}
Recent studies in RLVR highlight the critical role of token-level entropy. 
\citet{Yang2025DoNL} argue that low-probability tokens disproportionately influence model updates due to their large gradient magnitudes, which hinders effective learning of LMs. 
Other works observe that meaningful updates are concentrated at sparse ~`key tokens' representing pivotal decision junctions~\citep{Wang2025BeyondT8, arbuzov2025beyond}.
\citet{Wang2025EntropyBasedAW} further leverage these entropy patterns to design adaptive weighting strategies that prioritize uncertain transitions during post-training.

%% file: paper/conclusion.tex
\section{CONCLUSION}
This work investigates the coupling of pre-training and post-training, identifying the mid-training phase as a critical turning point. We introduce ReMiT, a method that leverages an RL reference  to assign dynamic, token-wise weights during mid-training, guiding the base model toward pivotal reasoning tokens. Unlike methods that discard data, ReMiT employs soft reweighting to preserve semantic coherence and training efficiency. Evaluations across multiple model families demonstrate that ReMiT consistently enhances both mid-training and post-training, establishing an iterative feedback loop between the base and the RL model.

%% file: paper/appendix.tex
\newpage
\appendix
\onecolumn

\section{Background: The Mid-Training Phase}\label{sec:app:bg}
\begin{figure}[ht]
    \centering
    \includegraphics[width=0.62\linewidth]{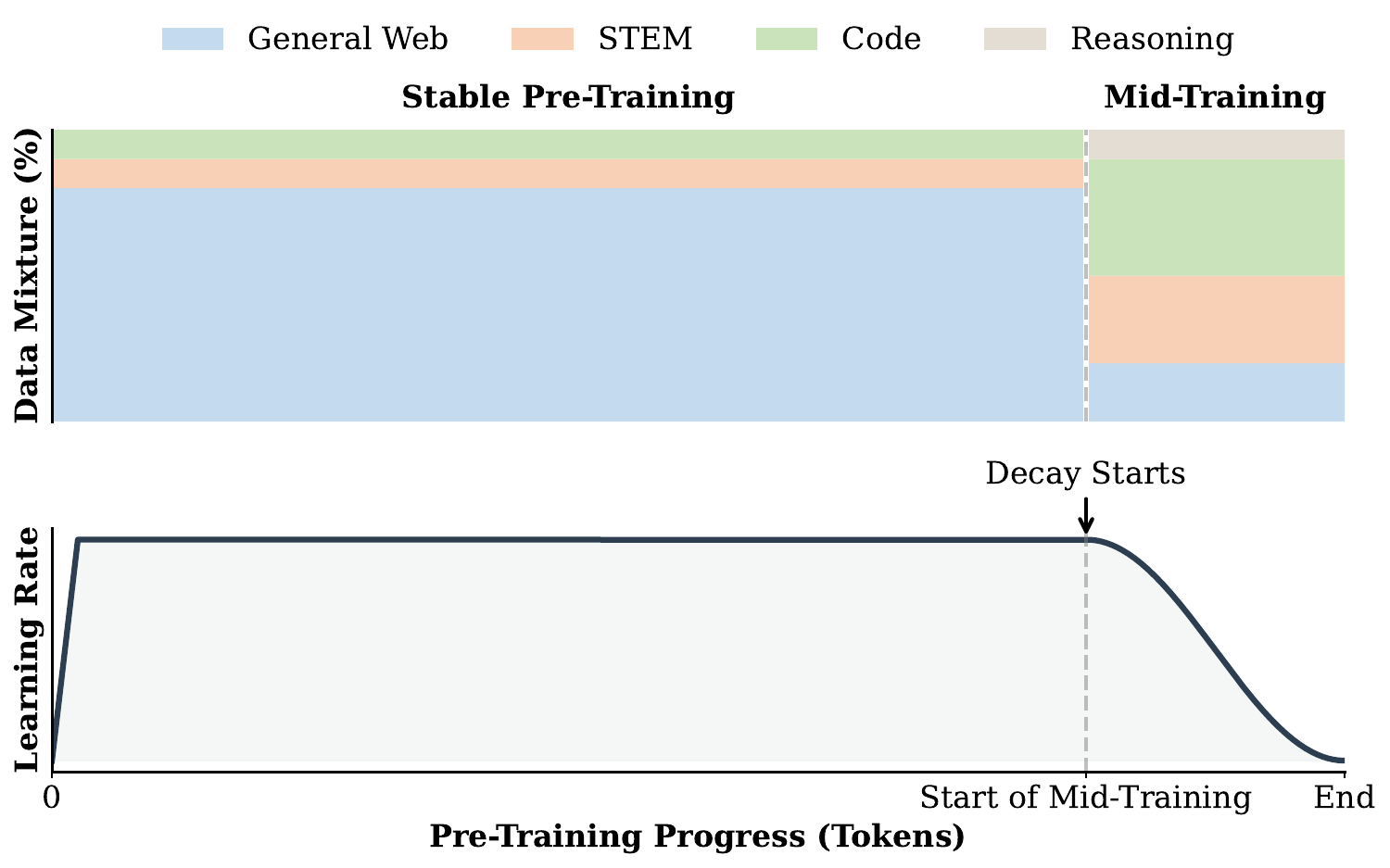}
    \caption{Overview of the pre-training pipeline. Top: During mid-training, the data distribution shifts from general web text to a higher-quality mixture enriched with code, STEM, and reasoning content. Bottom: The learning rate schedule is synchronized with this data shift, exhibiting rapid decay throughout the mid-training phase.}
    \label{fig:app:lr_schedule}
    \vspace{-3mm}
\end{figure}

Standard large language model (LLM) training typically comprises two stages: pre-training on large-scale corpora and post-training for alignment. Recent work highlights an intermediate stage---often termed \emph{mid-training} or \emph{annealing}---that bridges broad knowledge acquisition and the refinement of specialized capabilities. Mid-training is commonly associated with two shifts:

\vspace{-3mm}
\paragraph{High-quality data shift.}
Whereas stable pre-training relies primarily on large volumes of general web text to build broad coverage, mid-training emphasizes data quality over quantity. As illustrated in the top panel of Figure~\ref{fig:app:lr_schedule}, this phase increases the proportion of STEM, code, and reasoning-focused data, which can encourage the model to develop stronger competence in complex problem solving.

\vspace{-3mm}
\paragraph{Learning-rate decay.}
To incorporate this higher-quality data effectively, training typically applies an accelerated learning-rate decay. As shown in the bottom panel of Figure~\ref{fig:app:lr_schedule}, the learning rate decreases from the pre-training plateau to a near-zero value. This process can stabilize optimization and consolidate acquired behaviors, making mid-training a natural window for injecting or reinforcing reasoning-oriented capabilities before post-training.

\section{Mathematical Derivations}\label{sec:app:derivation}
\label{app:math}

In this section, we provide the rigorous derivations supporting the theoretical motivation in Section~\ref{sec:theory}.

\subsection{Decomposition of the ReMiT Objective}
\label{app:decomposition}

We show that minimizing the ReMiT objective is equivalent to minimizing the KL divergence between the implicit target $q_w$ and the model $\pi_\theta$.
Recall the ReMiT loss for a single step (ignoring the expectation over contexts for brevity):
\begin{equation}
    \mathcal{L}_{\text{ReMiT}} = - \sum_{x} p_{\text{data}}(x) w(x) \log \pi_\theta(x).
\end{equation}
Using the definition of the implicit target $q_w(x) = \frac{1}{Z_w} p_{\text{data}}(x) w(x)$, we can substitute $p_{\text{data}}(x) w(x) = Z_w q_w(x)$:
\begin{align}
    \mathcal{L}_{\text{ReMiT}} &= - \sum_{x} Z_w q_w(x) \log \pi_\theta(x) \\
    &= Z_w \left( - \sum_{x} q_w(x) \log \pi_\theta(x) \right).
\end{align}
Now, consider the KL divergence $D_{\text{KL}}(q_w \parallel \pi_\theta)$:
\begin{equation}
    D_{\text{KL}}(q_w \parallel \pi_\theta) = \sum_{x} q_w(x) \log \frac{q_w(x)}{\pi_\theta(x)} = \underbrace{\sum_{x} q_w(x) \log q_w(x)}_{-H(q_w)} - \sum_{x} q_w(x) \log \pi_\theta(x).
\end{equation}
Substituting the cross-entropy term back into the loss equation:
\begin{equation}
    \mathcal{L}_{\text{ReMiT}} = Z_w \left( D_{\text{KL}}(q_w \parallel \pi_\theta) + H(q_w) \right).
\end{equation}
Since the partition function $Z_w$ and the entropy $H(q_w)$ depend only on the fixed reference model and data distribution (independent of $\theta$), we have:
\begin{equation}
    \nabla_\theta \mathcal{L}_{\text{ReMiT}} = Z_w \nabla_\theta D_{\text{KL}}(q_w \parallel \pi_\theta).
\end{equation}
Note that while $Z_w$ acts as a scaling factor varying per context, the gradient direction aligns with minimizing the divergence to $q_w$.

\subsection{First-Order Expansion of KL Divergence}
\label{app:expansion}

We verify the condition under which an update towards $q_w$ reduces the divergence to the optimal policy $\pi^*$. Let $J(\pi) = D_{\text{KL}}(\pi^* \parallel \pi)$. We consider a perturbation of the policy $\pi$ in the direction of a target distribution $q$: $\pi' = \pi + \epsilon(q - \pi)$, where $\epsilon$ is a small learning rate.

The functional derivative (Fréchet derivative) of the KL divergence $J(\pi)$ with respect to $\pi(x)$ is:
\begin{equation}
    \frac{\delta J}{\delta \pi(x)} = - \frac{\pi^*(x)}{\pi(x)}.
\end{equation}
Using a first-order Taylor expansion, the change in divergence is:
\begin{align}
    J(\pi') - J(\pi) &\approx \sum_{x} \frac{\delta J}{\delta \pi(x)} \cdot (\pi'(x) - \pi(x)) \\
    &= \sum_{x} \left( - \frac{\pi^*(x)}{\pi(x)} \right) \cdot \epsilon (q(x) - \pi(x)) \\
    &= -\epsilon \sum_{x} \pi^*(x) \left( \frac{q(x) - \pi(x)}{\pi(x)} \right).
\end{align}
For the divergence to decrease (i.e., $J(\pi') < J(\pi)$), we require the change to be negative, which implies:
\begin{equation}
    \sum_{x} \pi^*(x) \frac{q(x) - \pi(x)}{\pi(x)} > 0.
\end{equation}
This confirms Equation~\eqref{eq:consistency_condition} in the main text.

\subsection{Relation Between ReMiT Updates and Distributional Motion}
\label{app:connection}
\begin{figure}[!t]
    \centering
    \includegraphics[width=0.99\linewidth]{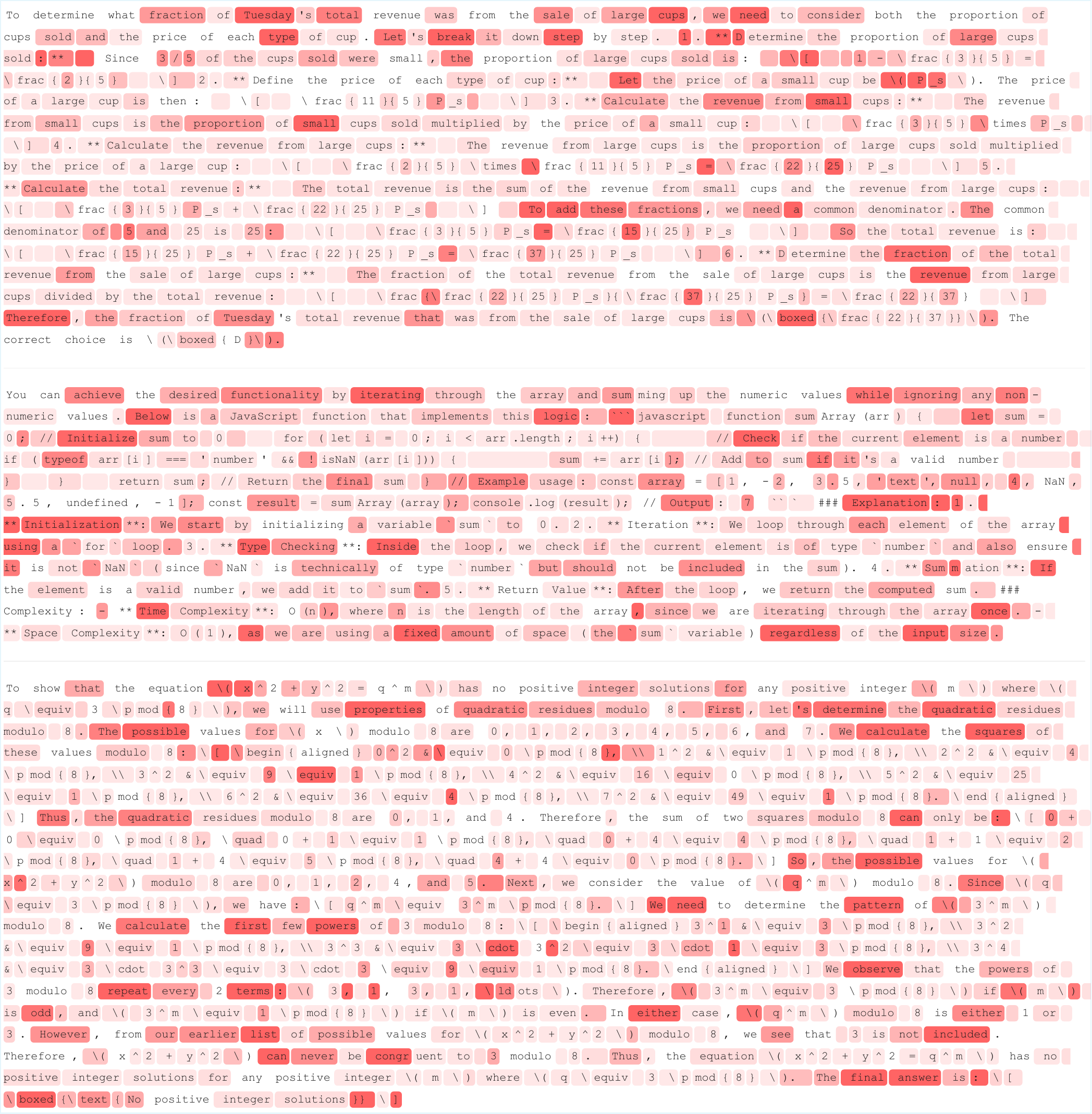}
    \caption{Token distribution divergence between the Base and RL models. Background intensity encodes the per-token log-probability margin ($\Delta\log p=\log p_{\mathrm{RL}}-\log p_{\mathrm{base}}$), with deeper red indicating higher confidence in the ground-truth token from the RL model.}
    \label{fig:app:visualization}
\end{figure}
In standard gradient descent, the parameter update $\theta_{new} \leftarrow \theta - \eta \nabla_\theta \mathcal{L}$ induces a change in the distribution space. For log-linear models (like the softmax output layer of LLMs), the natural gradient update direction in the probability simplex is proportional to the difference between the target and the current distribution:
\begin{equation}
    \Delta \pi(x) \propto q_w(x) - \pi_\theta(x).
\end{equation}
This justifies our assumption in Appendix A.2 that minimizing the ReMiT objective effectively moves the policy distribution $\pi_\theta$ in the direction of $q_w - \pi_\theta$. Thus, if the alignment condition (Equation~\eqref{eq:consistency_condition}) holds, the parameter update on $\theta$ translates to a functional improvement towards $\pi^*$.

\section{Visualization of Log-Probability Discrepancies Between Different Models}
\label{sec:app:visualize}
\begin{figure}[!t]
    \centering
    \includegraphics[width=0.9\linewidth]{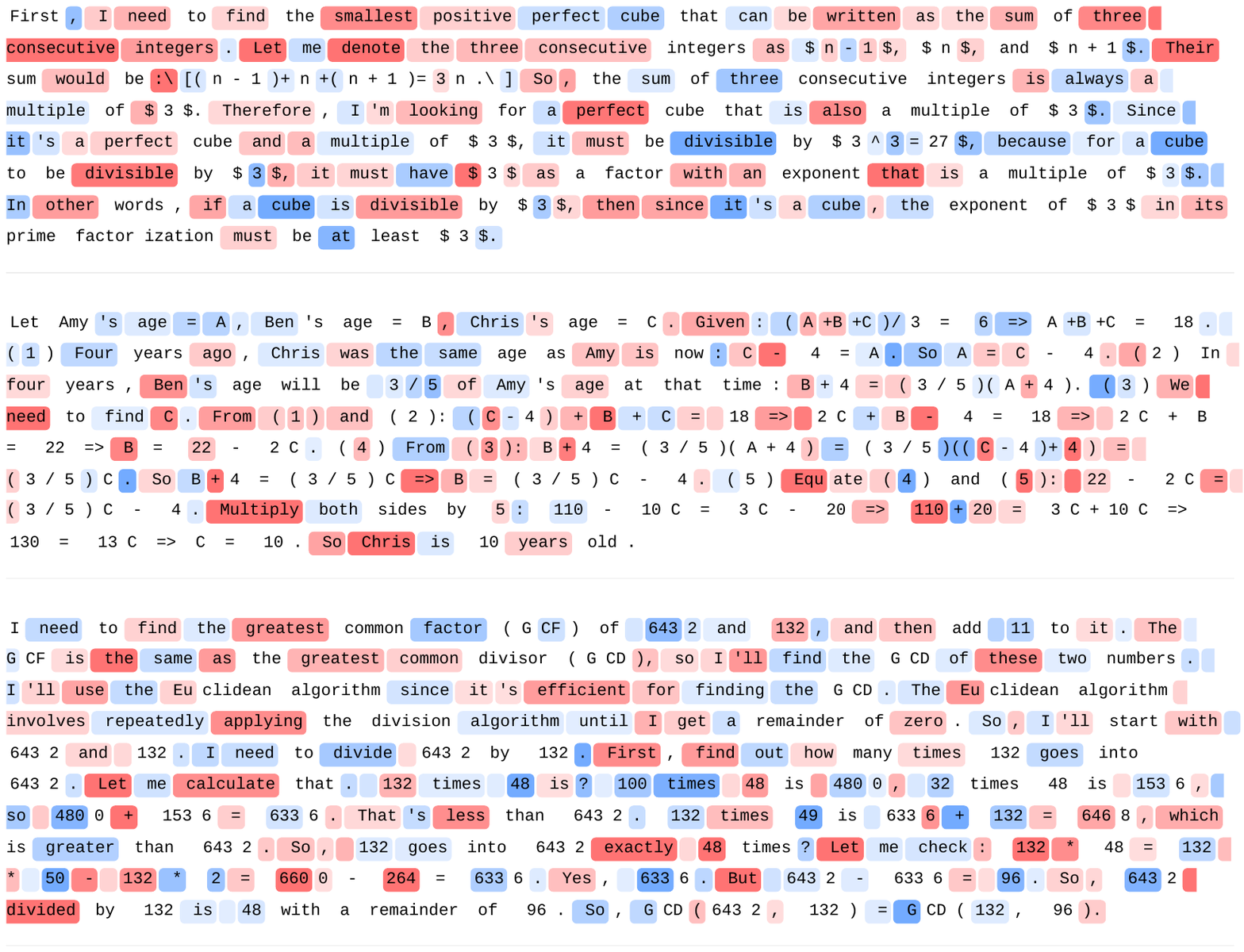}
    \caption{A comparison of token-level preferences between the RL and SFT models. The background color of each ground-truth token indicates the log-probability margin $\Delta \log p=\log p_{\mathrm{RL}}-\log p_{\mathrm{SFT}}$. Deeper red signifies a stronger preference from the RL model, deeper blue from the SFT model, and white denotes a negligible difference $(|\Delta \log p|<0.1)$.}
    \label{fig:app:sftrl}
\end{figure}

\subsection{Base Model versus RL Model}
To identify where the RL model surpasses the base model on a given sentence, Section~\ref{sec:bg2} presents an example visualizing log-probability discrepancies. 
Figure~\ref{fig:app:visualization} provides additional examples for further analysis. 
For each token, the background color encodes the RL-to-Base log-probability margin ($\Delta \log p$): lighter shades indicate smaller margins, whereas deeper red indicates that the RL model assigns higher likelihood to the ground-truth token. 
Most tokens show modest margins, while a small subset exhibits large positive margins (deep red). 
This pattern suggests that the RL model's superior reasoning is driven by a limited set of pivotal tokens, with the remainder largely following the base model's distribution.

\subsection{SFT Model versus RL Model}
We now analyze the rationale for selecting the RL model over the SFT model as a reference during the mid-training process. 
A key justification lies in their distinct token-level preferences, which we now examine in greater detail. 
As our examples illustrate, the RL model assigns a significantly higher likelihood to discourse connectives (e.g., `First', `But', `So', `Let'). 
This preference, visualized by a deeper red background for the corresponding tokens, is consistent with the enhanced reasoning capabilities cultivated during reinforcement learning, confirming the RL model as the more suitable reference for ReMiT.

\section{Experimental Configuration}\label{sec:app:eval}
\subsection{Mid-Training Corpus}
For the data setup, we utilize the official mid-training mixes associated with each model family. 
These corpora share a similar high-quality composition, blending general web text with domain-specific data such as code, mathematics, and STEM reasoning.
For all three models, the mid-training phase is conducted on a total of 50B tokens. The sequence length is set to 4,096 for OLMo and 8,192 for both SmolLM3 and Youtu-LLM, consistent with their original configurations.

\begin{figure}[!t]
    \centering
    \includegraphics[width=0.95\linewidth]{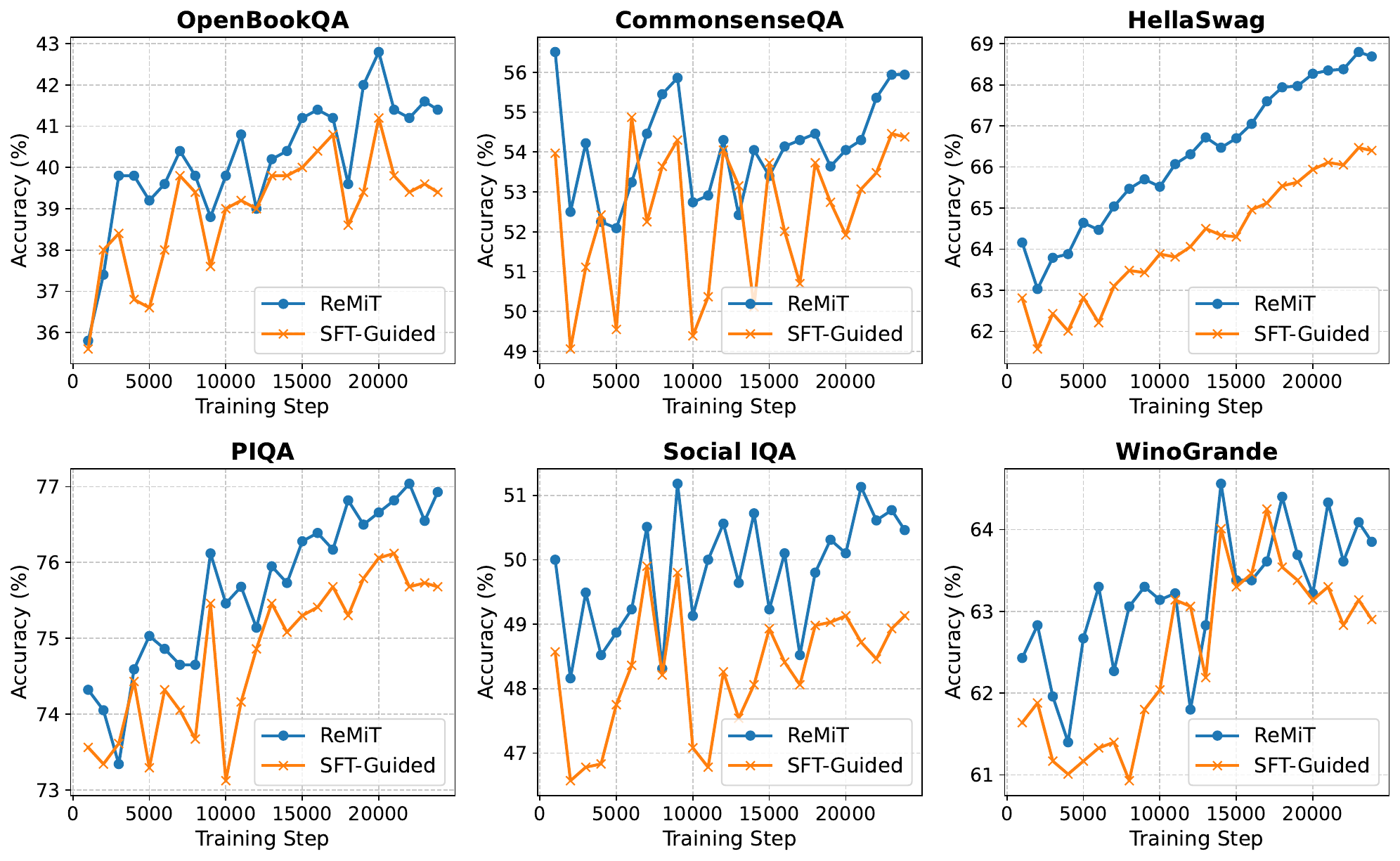}
    \caption{Comparison of downstream performance between SFT-Guided and ReMiT during mid-training on OLMo-1B. The results demonstrate that our RL-guided approach ReMiT consistently outperforms the SFT-guided baseline throughout the training trajectory.}
    \label{fig:app:sga}
\end{figure}

\subsection{Optimization Hyperparameters}
We employ a cosine decay schedule down to a minimum learning rate of 1e-7. The specific configurations are: \begin{itemize} \item \textbf{OLMo-1B:} Maximum learning rate 3e-4, Global batch size 2M tokens. \item \textbf{SmolLM3-3B:} Maximum learning rate 2e-4, Global batch size 2M tokens. \item \textbf{Youtu-LLM-2B:} Maximum learning rate 2e-4, Global batch size 64M tokens. \end{itemize} These settings are consistent with the configurations used during the respective stable pre-training phases of each model.

\subsection{Evaluation of Pre-trained Base LMs}
We comprehensively evaluate the pre-trained base models by measuring their few-shot performance across a suite of widely used downstream benchmarks, including: GSM8K (\citet{cobbe2021training}; 8-shot), Minerva Math (\citet{lewkowycz2022solving}; 4-shot), GPQA (\citet{rein2024gpqa}; 5-shot), BigBenchHard (BBH; \citet{suzgun2022challenging}; 3-shot CoT), IFEval (IFE; \citet{zhou2023instruction} 3-shot), HumanEval (HE; \citet{zheng2023codegeex}; 0-shot), MBPP (\citet{austin2021program}; 3-shot), TruthfulQA (TQA; \citet{lin2021truthfulqa}; 3-shot), ARC-Challenge (\citet{clark2018think}; 25-shot), and $\text{MMLU}_{\text{Pro}}$ (\cite{wang2024mmlu}; 5-shot).
Unless otherwise specified, we report Pass@1 for both MBPP and HumanEval.
In extended experiments, we also evaluate additional benchmarks such as HellaSwag~\citep{zellers2019hellaswag}, PIQA~\citep{bisk2020piqa}, Social IQA~\cite{sap2019socialiqa}, WinoGrande~\cite{sakaguchi2021winogrande}, OpenBookQA~\cite{mihaylov2018can}, CommensenseQA~\cite{talmor2019commonsenseqa} and MuSR~\citep{Sprague2023MuSRTT}.

\subsection{Evaluation of Post-trained LMs}
Given that the post-trained models already exhibit instruction-following ability, we evaluate their zero-shot accuracy. We report results on 10 widely used benchmarks: GSM8K~\citep{cobbe2021training}, Minerva Math~\citep{lewkowycz2022solving}, 
MBPP~\citep{austin2021program}, MBPP+~\citep{evalplus}, HumanEval (HE; \citet{zheng2023codegeex}), HumanEval+ (HE+; \citet{evalplus}), ARC-Challenge~\citep{clark2018think}, $\text{GPQA}_{\text{Diamond}}$ ($\text{GPQA}_{\text{DM}}$; \citet{rein2024gpqa}), $\text{MMLU}_{\text{Pro}}$~\citep{wang2024mmlu}, IFEval (IFE; \citet{zhou2023instruction}).

\section{Extended Analysis and Results}\label{sec:app:exp}
\input{table/exp3}
\input{table/exp2}
\input{table/exp4}
\begin{figure}[!t]
    \centering
    \includegraphics[width=0.88\linewidth]{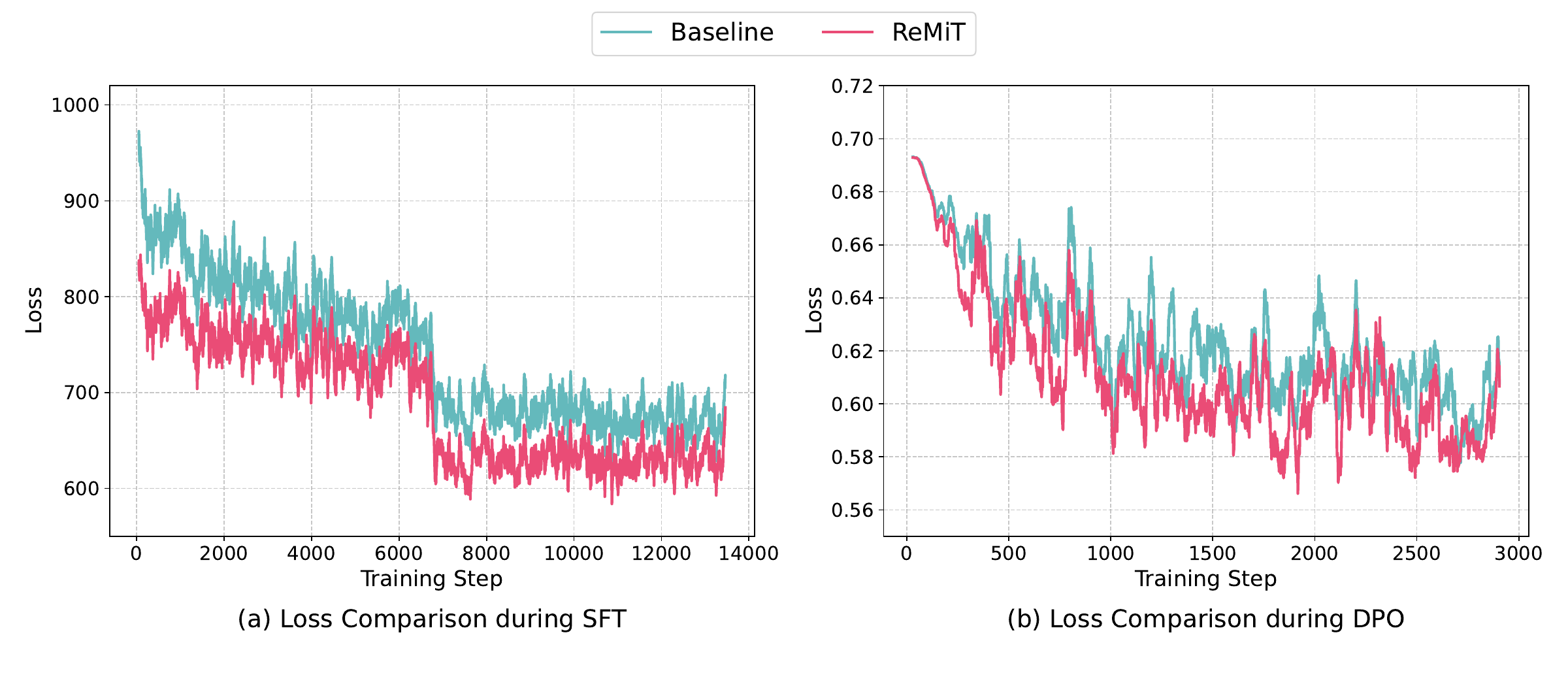}
    \caption{Loss trajectories during SFT and DPO on OLMo-1B.}
    \label{fig:app:post_training_loss}
    \vspace{-4mm}
\end{figure}
\begin{figure}[!t]
    \centering
    \includegraphics[width=0.85\linewidth]{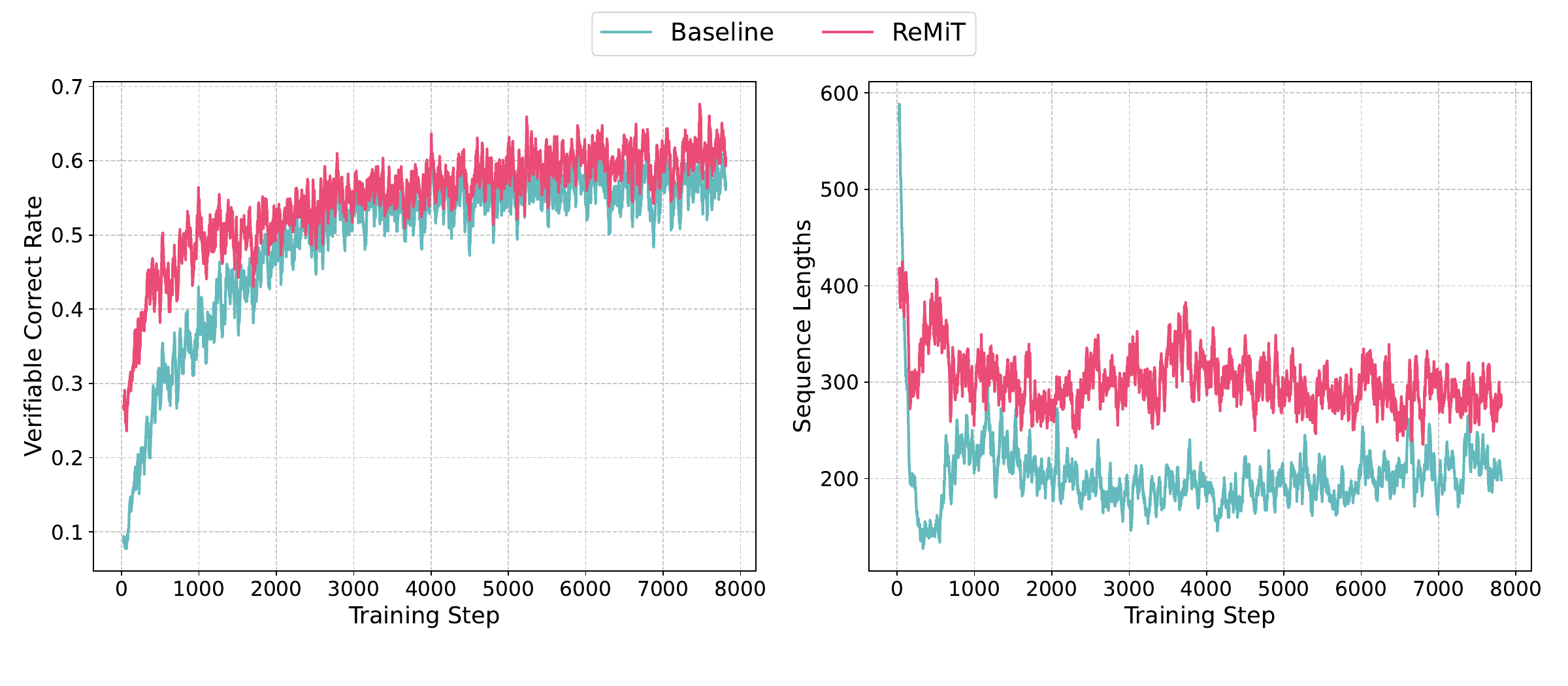}
    \caption{Performance comparison at the reinforcement learning stage on OLMo-1B.}
    \label{fig:app:rl_zero}
    \vspace{-4mm}
\end{figure}
\subsection{Pre-Training Results}
In this section, we provide extended experimental results comparing different mid-training methodologies, with a particular focus on the ablation variants discussed in Section~\ref{sec:ablation}.

\minisection{Reference Policy Ablation: RL vs. SFT}
We first present a detailed comparison between utilizing the RL model (ReMiT) versus the SFT model (SFT-Guided) as the reference policy. 
Figure~\ref{fig:app:sga} illustrates the downstream performance trajectories on OLMo-1B throughout the mid-training process. 
It is evident that ReMiT consistently outperforms SFT-Guided across the training steps. 
This superiority is further corroborated by the comprehensive benchmark results in Table~\ref{tab:exp3}, empirically validating the choice of the RL model as the optimal reference for guiding the base model.
For analysis investigating why these models yield different results, we provide visualizations of their distinct token-level preferences in Appendix~\ref{sec:app:visualize}.

\minisection{Effectiveness of Knowledge Distillation}

Echoing the results in Figure~\ref{fig:radar}(a) of the main text, we observe that knowledge distillation (KD) achieves impressive performance gains immediately in the mid-training phase. 
This demonstrates the effectiveness of strict distribution matching for transferring capabilities in the short term.
However, a critical limitation emerges upon further evaluation: these advantages fail to sustain after subsequent post-training. 
This phenomenon suggests that while KD effectively clones the teacher's current distribution, it may compromise the model's plasticity for future adaptation, which we will show in the following section regarding post-training results.

\subsection{Post-Training Results}
To further demonstrate that ReMiT enhances performance beyond the mid-training stage, we conduct additional post-training on the mid-trained base models. 
As illustrated in Figure~\ref{fig:app:post_training_loss}, the loss trajectories for both SFT and DPO stages show that ReMiT consistently achieves lower loss values compared to the standard approach. 
Furthermore, Figure~\ref{fig:app:rl_zero} presents the training dynamics during the RLVR stage, indicating that ReMiT achieves a higher verifiable correct rate while also maintaining longer sequence lengths.

Table~\ref{tab:exp2} details the post-trained performance across 10 downstream benchmarks. 
The results confirm that the gains acquired during mid-training are effectively preserved throughout the entire post-training pipeline. 
Notably, compared to KD, whose early advantages tend to diminish during subsequent alignment steps due to over-fitting the teacher's distribution, ReMiT demonstrates sustained superiority. 
This observation aligns with the conclusions drawn from Figure~\ref{fig:exp3} (b) and~\ref{fig:radar}(b) in the main text, highlighting ReMiT's robust generalization capability.

\section{Discussion}\label{sec:app:discuss}
\begin{figure}[!t]
    \centering
    \includegraphics[width=0.9\linewidth]{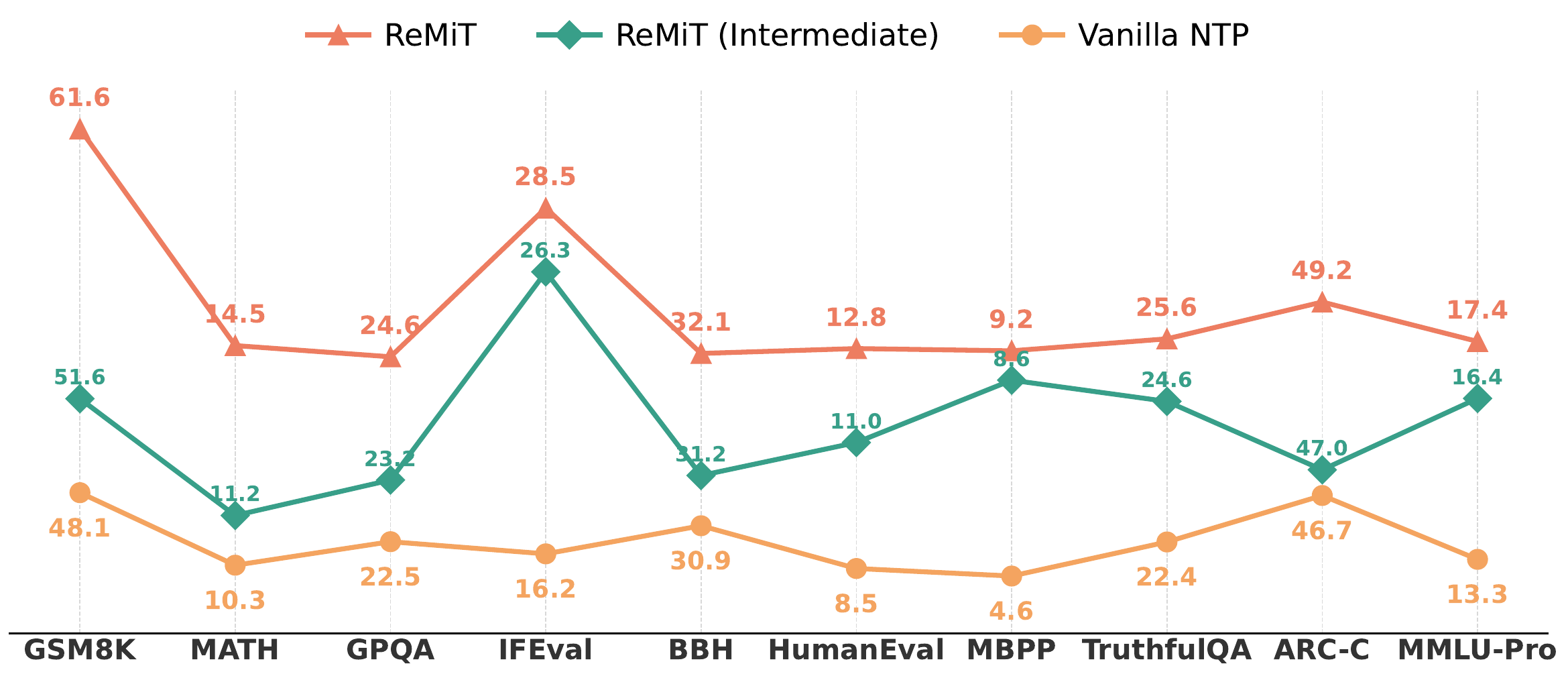}
    \caption{Impact of RL reference quality on mid-training performance across 10 downstream tasks. We compare ReMiT (using a fully converged RL model) against ReMiT (Intermediate) which uses a checkpoint trained for only 10\% of RL steps.}
    \label{fig:app:weaker_rl}
\end{figure}
\subsection{Impact of RL Reference Quality}
We study how the quality of the RL reference affects mid-training by using an intermediate RL checkpoint (trained for only 10\% of the total RL steps) as a weaker reference, denoted ReMiT (Intermediate). We compare this variant with standard ReMiT (using the fully converged RL model) and the Vanilla NTP baseline. Results are shown in Figure~\ref{fig:app:weaker_rl}.

\vspace{-5mm}
\paragraph{Sensitivity to signal quality.}
We observe a clear performance ordering: ReMiT $>$ ReMiT (Intermediate) $>$ Vanilla NTP. The observed positive correlation between reference quality and downstream performance confirms that ReMiT effectively leverages the specific reasoning priors embedded in the reference signal.

\vspace{-5mm}
\paragraph{Robustness and safety boundaries.}
Despite being trained for only 10\% of the RL steps, ReMiT (Intermediate) consistently outperforms Vanilla NTP, indicating robustness to imperfect guidance. Moreover, the clipping mechanism in Equation~(5) provides an explicit safeguard by limiting extreme token weights, thereby reducing the risk of overfitting to noise or bias in a weaker reference.

\vspace{-5mm}
\paragraph{Implications for the in-pipeline strategy.}
These findings further support the in-pipeline design (Figure~\ref{fig:pipeline}): the observed gains can be obtained using an RL reference produced within the same pipeline and of the same parameter scale, yielding a reliable and consistent signal without relying on larger, external teachers. This reinforces our central contribution---an iterative evolution cycle between pre-training and post-training. By feeding post-training improvements back into mid-training, ReMiT forms a self-reinforcing flywheel that enables continual progress while avoiding the additional computational cost of training separate, external reference models.

\subsection{Computational Efficiency and Overhead Analysis}
\begin{figure}[!t]
    \centering
    \includegraphics[width=0.88\linewidth]{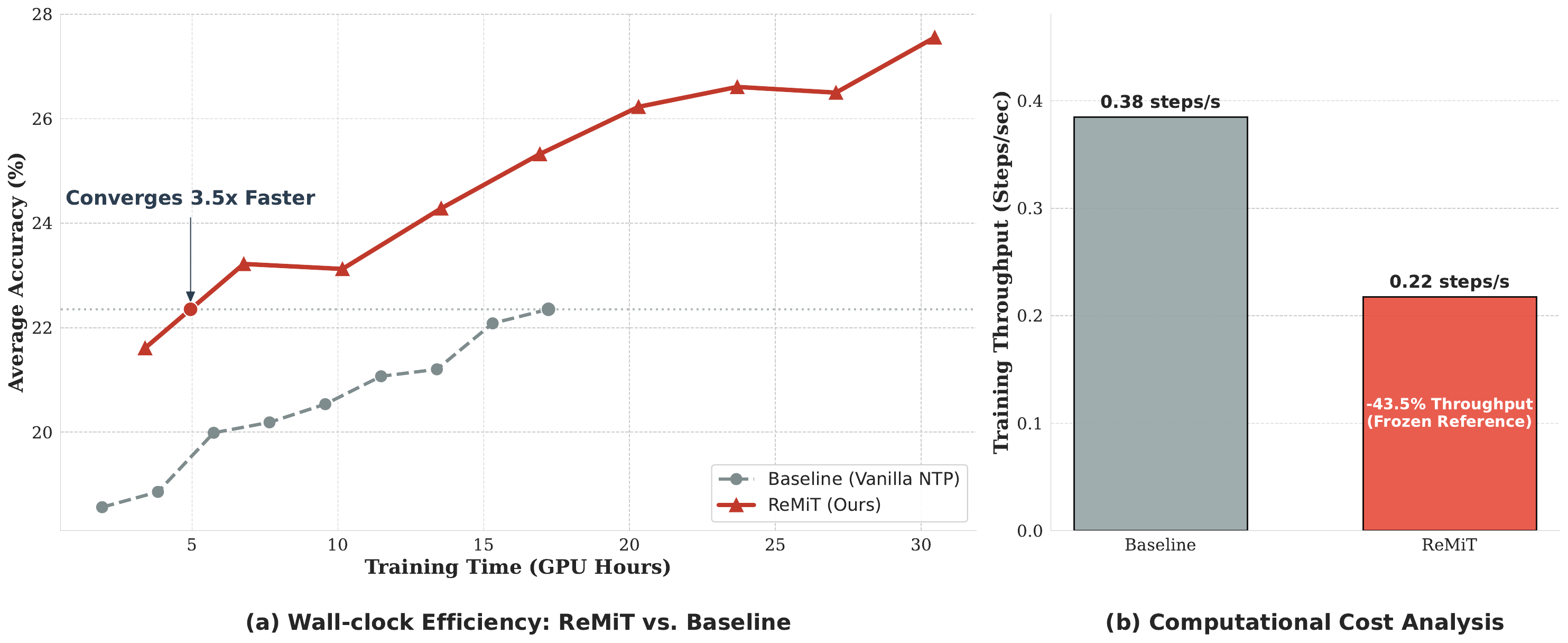}
    \caption{Training efficiency analysis of OLMo-1B on the same 50B mid-training corpus. (a) Wall-clock Convergence: Despite the computational overhead of the reference model, ReMiT demonstrates superior efficiency, converging 3.3x faster than the baseline in terms of GPU hours. (b) Throughput Comparison: While ReMiT incurs a reduction in training throughput (steps/sec) due to the additional forward pass, this cost is effectively amortized by its rapid convergence rate.}
    \label{fig:app:compt_overhead}
\end{figure}
A potential concern with reference-guided approaches is the computational overhead introduced by the additional forward pass of the reference model. To address this, we analyze the trade-off between training throughput and convergence speed, demonstrating that ReMiT offers a superior compute-performance ratio compared to standard next token prediction.

\vspace{-5mm}
\paragraph{Time-to-Convergence Efficiency.}
While Figure~\ref{fig:abstract} establishes ReMiT’s sample efficiency—showing it reaches baseline performance using significantly fewer tokens ($6\times$ faster in token count) —Figure 15 provides a critical wall-clock analysis that accounts for real-world computational costs. As shown in Figure 15(a), ReMiT achieves a 3.5x speedup in time-to-convergence compared to the Vanilla NTP baseline, measured in GPU hours. This result is significant because it confirms that the gains in sample efficiency vastly outweigh the reduction in training throughput.

\vspace{-5mm}
\paragraph{Overhead vs. Gain.}
Figure 15(b) explicitly quantifies the computational cost: ReMiT incurs a $43\%$ reduction in throughput (steps/sec) due to the inference overhead of the frozen RL reference. However, this cost is effectively amortized by the method's rapid learning trajectory. Crucially, the performance ceiling achieved by ReMiT in Figure 15(a) is not merely a result of accelerated convergence; the curve shows that ReMiT reaches accuracy levels that the Vanilla NTP baseline struggles to attain even with extended training. This indicates that ReMiT provides a qualitative improvement in reasoning capabilities that cannot be replicated simply by training a standard model for longer periods or with more tokens.
Moreover, given that the mid-training stage typically occupies a short window at the end of pre-training, this temporary overhead does not pose a barrier for scaling to larger models.
Future research could investigate strategies to mitigate this overhead, for instance, through offline weight computation or asynchronous scoring mechanisms.

%% file: table/exp3.tex
\begin{table*}[!t]
\renewcommand\arraystretch{1.02}
\setlength{\tabcolsep}{6pt}
\centering
\caption{Extended evaluation of pre-training models across 10 widely used downstream tasks. The best scores within each model family are \textbf{boldfaced}.}
\resizebox{0.99\textwidth}{!}{
\footnotesize
\begin{tabular}{c|cccccccccc|c}
\toprule
\textbf{Method} & \textbf{GSM8K} & \textbf{MATH} & \textbf{GPQA} & \textbf{BBH} & \textbf{IFE} & \textbf{HE} & \textbf{MBPP} & \textbf{TQA} & \textbf{ARC-C} & \textbf{MMLU\textsubscript{P}} & \textbf{Avg.} \\
\midrule
\rowcolor{lightpink}\multicolumn{12}{c}{Mid-Training on OLMo-1B} \\
\midrule
Vanilla NTP & 48.14 & 10.26 & 22.54 & 30.87 & 16.19 & 8.54 & 4.60 & 22.40 & 46.67 & 13.31 & 22.35 \\
SFT-Guided & 50.34 & 12.16 & 23.44 & 31.15 & 13.07 & 10.98 & 9.00 & 24.11 & 47.70 & 15.23 & 23.72 \\
KD & 61.26 & \textbf{15.88} & 23.88 & \textbf{32.11} & \textbf{45.56} & \textbf{13.41} & \textbf{13.20} & \textbf{25.83} & \textbf{49.23} & \textbf{17.45} & \textbf{29.78} \\
ReMiT & \textbf{61.64} & 14.50 & \textbf{24.55} & 32.07 & 28.54 & 12.80 & 9.20 & 25.58 & \textbf{49.23} & 17.44 & 27.56\\
\midrule
\rowcolor{lightpink}\multicolumn{12}{c}{Mid-Training on Youtu-LLM-2B}
\\
\midrule
Vanilla NTP & 49.51 & \textbf{25.00} & 27.90 & 44.37 & 32.61 & 37.20 & 46.60 & 26.56 & 53.33 & 24.93 & 36.80\\
SFT-Guided & 47.54 & 19.84 & \textbf{30.13} & 44.60 & 32.73 & 32.93 & 43.40 & 25.46 & 52.65 & 24.42 & 35.37 \\
KD & \textbf{60.42} & 23.86 & 25.67 & \textbf{50.71} & \textbf{38.61} & 32.93 & 45.20 & \textbf{29.38} & 52.13 & \textbf{27.77} & \textbf{38.67} \\
ReMiT & 52.69 &  24.50 & 29.69 & 47.21 & 36.93 & \textbf{39.94} & \textbf{47.00} & 27.42 & \textbf{54.61} & 25.85 & 38.58\\
\bottomrule
\end{tabular}
}
\label{tab:exp3}
\end{table*}

%% file: table/exp2.tex
\begin{table*}[!t]
\renewcommand\arraystretch{1.02}
\setlength{\tabcolsep}{6pt}
\centering
\caption{Zero-shot accuracy of post-trained models on OLMo-1B across 10 widely used downstream tasks. The best scores are \textbf{boldfaced}.}
\resizebox{0.99\textwidth}{!}{
\footnotesize
\begin{tabular}{c|cccccccccc|c}
\toprule
\textbf{Method} & \textbf{GSM8K} & \textbf{MATH} &  \textbf{MBPP} & \textbf{MBPP+} & \textbf{HE} & \textbf{HE+} &\textbf{ARC-C} & \textbf{GPQA\textsubscript{DM}} & \textbf{MMLU\textsubscript{P}} & \textbf{IFE} & \textbf{Avg.} \\
\midrule
\rowcolor{lightpink}\multicolumn{12}{c}{Stage: SFT} \\
\midrule
Vanilla NTP & 42.61 & 12.28  & 16.20 & 25.66 & 21.95 & 18.90 & 40.96 & 24.75 & 14.34 & 56.83 & 27.45\\
KD & 41.55 & 15.22 & \textbf{21.20} & \textbf{33.33} & \textbf{30.49} & \textbf{26.22} & 38.91 & 25.25 & \textbf{15.41} & 58.03 & \textbf{30.56}\\
ReMiT & \textbf{47.00} &  \textbf{16.54} & 19.60 & 29.10 & 26.22 & 21.95 & \textbf{41.38} & \textbf{25.76} & 13.56 & \textbf{59.11} & 30.02\\
\midrule
\rowcolor{lightpink}\multicolumn{12}{c}{Stage: DPO}
\\
\midrule
Vanilla NTP & \textbf{57.70} & 15.10  & 9.20 & 18.78 & 24.39 & 21.95 & 43.43 & 21.72 & 14.61 & 67.39 & 29.43\\
KD & 49.66 & 15.96 & \textbf{12.80} & \textbf{22.75} & 22.56 & 20.73 & 39.25 & \textbf{26.34} & \textbf{15.80} & 67.87 & 29.37\\
ReMiT & 55.04 & \textbf{17.96} & 12.40 & 21.69 & \textbf{25.61} & \textbf{22.56} & \textbf{45.22} & 23.23 & 15.78 & \textbf{69.18} & \textbf{30.87}\\
\midrule
\rowcolor{lightpink}\multicolumn{12}{c}{Stage: RL}
\\
\midrule
Vanilla NTP & \textbf{66.72} & 18.60 & 8.80 & 17.33 & 19.51 & 17.07 & 43.00 & 20.20  & 14.58 & 69.90 & 29.57\\
KD & 62.77 & 16.84  & 11.50 & 21.30 & 24.39 & 22.56 & 41.38 & 24.75 & 15.24 & 69.30 & 31.00\\
ReMiT & 65.73 & \textbf{19.66}  & \textbf{11.60} & \textbf{22.09} & \textbf{28.35} & \textbf{24.39} & \textbf{44.80} & \textbf{25.25} & \textbf{16.73} & \textbf{70.62} & \textbf{32.92}\\
\bottomrule
\end{tabular}
}
\label{tab:exp2}
\end{table*}

%% file: table/exp4.tex
\begin{table*}[!t]
\renewcommand\arraystretch{1.02}
\setlength{\tabcolsep}{6pt}
\centering
\caption{Zero-shot accuracy of post-trained models on Youtu-LLM-2B across 10 widely used downstream tasks. The best scores are \textbf{boldfaced}.}
\resizebox{0.99\textwidth}{!}{
\footnotesize
\begin{tabular}{c|cccccccccc|c}
\toprule
\textbf{Method} & \textbf{GSM8K} & \textbf{MATH} &  \textbf{MBPP} & \textbf{MBPP+} & \textbf{HE} & \textbf{HE+} &\textbf{ARC-C} & \textbf{GPQA\textsubscript{DM}} & \textbf{MMLU\textsubscript{P}} & \textbf{IFE} & \textbf{Avg.} \\
\midrule
\rowcolor{lightpink}\multicolumn{12}{c}{Stage: SFT} \\
\midrule
Vanilla NTP & 57.32 & 22.46 & 46.60 & 62.17 & 56.71 & 50.61 & 43.77 & 24.75 & 23.66 & 65.83 & 45.39\\
KD & 63.38 & \textbf{27.28} & 45.00 & 61.90 & \textbf{57.32} & 51.22 & 43.00 & 27.27 & \textbf{27.02} & \textbf{66.55} & 46.99\\
ReMiT & \textbf{64.06} & 22.84 & \textbf{49.40} & \textbf{65.34} & \textbf{57.32} & \textbf{53.66} & \textbf{45.22} & \textbf{29.80} & 25.41 & 65.59 & \textbf{47.86}\\
\midrule
\rowcolor{lightpink}\multicolumn{12}{c}{Stage: DPO}
\\
\midrule
Vanilla NTP & 64.75 & 26.36 & \textbf{47.60} & 62.70 & 60.37 & 54.88 & 46.50 & 27.78 & 27.32 & 74.82 & 49.31\\
KD & \textbf{72.48} & \textbf{30.76} & 47.00 & 65.08 & 60.98 & 56.10 & 45.22 & 25.76 & \textbf{30.40} & 74.58 & 50.84\\
ReMiT & 71.34 & 28.62 & 47.00 & \textbf{67.46} & \textbf{64.02} & \textbf{59.15} & \textbf{47.10} & \textbf{30.30} & 29.40 & \textbf{75.54} & \textbf{51.99}\\
\bottomrule
\end{tabular}
}
\label{tab:exp4}
\end{table*}